\documentclass[sigconf]{acmart}
\usepackage{multirow}
\usepackage{colortbl}
\usepackage{pifont}
\usepackage{bbding}
\usepackage{tablefootnote}
\newcommand{\gr}{\rowcolor[gray]{.95}}
\UseRawInputEncoding
\AtBeginDocument{%
  }

\copyrightyear{2023}
\acmYear{2023}
\setcopyright{acmlicensed}
\acmConference[MM '23] {Proceedings of the 31st ACM International Conference on Multimedia}{October 29--November 3, 2023}{Ottawa, ON, Canada.}
\acmBooktitle{Proceedings of the 31st ACM International Conference on Multimedia (MM '23), October 29--November 3, 2023, Ottawa, ON, Canada}
\acmPrice{15.00}
\acmISBN{979-8-4007-0108-5/23/10}
\acmDOI{10.1145/3581783.3611744}
\begin{document}

\title{NightHazeFormer: Single Nighttime Haze Removal Using Prior Query Transformer}



 \author{Yun Liu}
 \affiliation{%
   \institution{College of Artificial Intelligence\\ Southwest University}
   \city{Chongqing}
   \country{China}
 }
\email{yunliu@swu.edu.cn}

  \author{Zhongsheng Yan}
 \affiliation{%
   \institution{College of Artificial Intelligence\\ Southwest University}
   \city{Chongqing}
   \country{China}
 }
\email{zhongshengyanzy@foxmail.com}

  \author{Sixiang Chen}
 \authornote{Corresponding author}
 \affiliation{%
   \institution{The Hong Kong University of Science and Technology (Guangzhou)}
   \city{Guangzhou}
   \country{China}
 }
\email{sixiangchen@hkust-gz.edu.cn}

 \author{Tian Ye}
 \authornotemark[1]
 \affiliation{%
   \institution{The Hong Kong University of Science and Technology (Guangzhou)}
   \city{Guangzhou}
   \country{China}
 }
 \email{owentianye@hkust-gz.edu.cn}

 \author{Wenqi Ren}
 \affiliation{%
   \institution{School of Cyber Science and Technology \\ Sun Yat-sen University}
   \city{Shenzhen Campus, Shenzhen}
   \country{China}
 }
 \email{renwq3@mail.sysu.edu.cn}

   \author{Erkang Chen}
 \affiliation{%
   \institution{School of Ocean Information Engineering \\ Jimei University}
   \city{Xiamen}
   \country{China}
  %
 }
 \email{ekchen@jmu.edu.cn}

\begin{abstract}
Nighttime image dehazing is a challenging task due to the presence of multiple types of adverse degrading effects including glow, haze, blur, noise, color distortion, and so on. However, most previous studies mainly focus on daytime image dehazing or partial degradations presented in nighttime hazy scenes, which may lead to unsatisfactory restoration results. In this paper, we propose an end-to-end transformer-based framework for nighttime haze removal, called NightHazeFormer. Our proposed approach consists of two stages: supervised pre-training and semi-supervised fine-tuning. During the pre-training stage, we introduce two powerful priors into the transformer decoder to generate the non-learnable prior queries, which guide the model to extract specific degradations. For the fine-tuning, we combine the generated pseudo ground truths with input real-world nighttime hazy images as paired images and feed into the synthetic domain to fine-tune the pre-trained model. This semi-supervised fine-tuning paradigm helps improve the generalization to real domain. In addition, we also propose a large-scale synthetic dataset called UNREAL-NH, to simulate the real-world nighttime haze scenarios comprehensively. Extensive experiments on several synthetic and real-world datasets demonstrate the superiority of our NightHazeFormer over state-of-the-art nighttime haze removal methods in terms of both visually and quantitatively. Dataset will be available at https://github.com/Owen718/NightHazeFormer.
\end{abstract}




\maketitle
\section{Introduction}
Under real-world nighttime haze imaging conditions, the illumination is dominated by various artificial light sources such as neon lights and they have different locations and colors with limited luminance range. Therefore, apart from the haze, the acquired degraded images also will be affected by multiple scattering, uneven illumination, glow, blur, hidden noise, etc. Compared to daytime image dehazing, how to recover clear images from nighttime hazy scenarios becomes a new challenging task.
\begin{figure}
  \vspace{0.5em}
  \setlength{\abovecaptionskip}{5pt}
  \setlength{\belowcaptionskip}{-17pt}
  \centering
  \includegraphics[width=\linewidth]{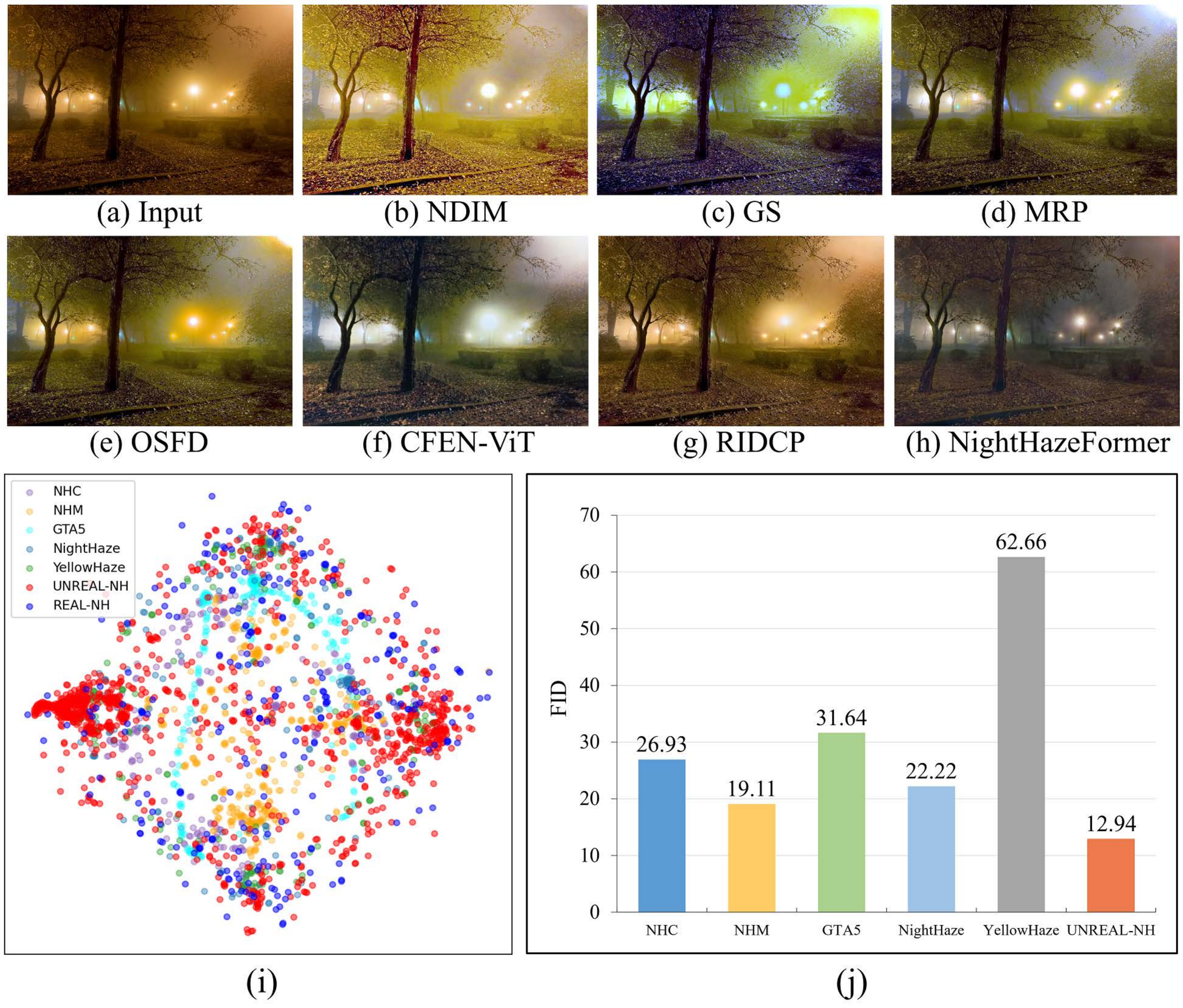}\\
  \caption{(a) Nighttime hazy image. (b)-(h) are the dehazed results with NDIM~\cite{zhang2014nighttime}, GS~\cite{li2015nighttime}, MRP~\cite{zhang2017fast}, OSFD~\cite{zhang2020nighttime}, CFEN-ViT~\cite{zhao2021complementary}, RIDCP~\cite{wu2023ridcp} and our NightHazeFormer. (i) The t-SNE map of various synthetic nighttime hazy image datasets. The feature distribution of degradations in our UNREAL-NH dataset is more closer to real-world nighttime hazy scenes (REAL-NH) than existing synthetic datasets. (j) Histogram of FID values for various synthetic datasets. Obviously, our UNREAL-NH exhibits the smallest FID value among all synthetic datasets, which quantitatively proves that our UNREAL-NH dataset is more realistic.
  }\label{comparisons1}
\end{figure}

To the best of our knowledge, significant progress has been made for daytime image dehazing. The current daytime haze removal approaches, including prior-based methods~\cite{he2010single,fattal2014dehazing,zhu2015fast,berman2018single,bui2018single,ju2019idgcp,ju2021idrlp,zhao2021single} and learning-based algorithms~\cite{ren2016single,cai,li2017aod,qin2020ffa,wu2021contrastive,chen2021psd,ye2022perceiving,chai2022pdd,yu2022source,chen2023dehrformer}, have limited effectiveness in restoring nighttime hazy images due to two reasons. First, the widely used haze imaging model~\cite{mccartney1976optics} is unable to fully describe the complex formation of a nighttime hazy image. Second, there exist notable degradation discrepancies between daytime and nighttime hazy scenarios, further hindering the recovery of nighttime hazy images.

To address the diverse types of degradations that occur in nighttime hazy environments, some new imaging models~\cite{zhang2014nighttime,li2015nighttime,liu2021single} are proposed to illustrate the degradation characteristics such as non-uniform illumination and glow. Subsequently, several model-based nighttime haze removal algorithms~\cite{zhang2014nighttime,li2015nighttime,zhang2017fast,zhang2020nighttime,liu2021single,wang2022variational,liu2022single,liu2022nighttime,wang2022rapid,liu2023multi} have been developed to restore the degraded images. Although the above approaches have achieved decent dehazing results to some extent, they cannot simultaneously overcome all types of degradations due to their focus on only partial corruption factors.

Compared to model-based methods, deep learning based haze removal networks~\cite{liao2018hdp,koo2019nighttime,yan2020nighttime,zhang2020nighttime,kuanar2022multi,zhao2021complementary,wu2023ridcp,jin2023enhancing} for nighttime hazy images are still limited, primarily due to the lack of realistic synthetic datasets. Existing large-scale synthetic datasets, such as NHC~\cite{zhang2020nighttime}, NHM~\cite{zhang2020nighttime}, GTA5~\cite{yan2020nighttime}, NightHaze~\cite{liao2018hdp} and YellowHaze~\cite{liao2018hdp}, are unable to comprehensively simulate the complex degradations presented in real-world nighttime hazy images, especially for light effects and spatially variant illumination. Consequently, the dehazing networks trained on these synthetic datasets usually suffer from poor generalization to real-world nighttime hazy images, leading to unsatisfactory restoration results. Additionally, most nighttime haze removal networks solely rely on synthetic datasets for training, making it challenging to acquire the domain knowledge from real data due to the domain shift problem.

To address the above issues, we develop~\textbf{NightHazeFormer}, a transformer-based network for nighttime haze removal, which consists of a supervised pre-training phase and a semi-supervised fine-tuning phase. For pre-training, we introduce two powerful physical priors, dark channel prior (DCP)~\cite{he2010single} and bright channel prior (BCP)~\cite{wang2013automatic}, into the transformer decoder to generate the non-learnable prior queries. These queries are served as the explicit degradations guidance for the self-attention transformer block and help our model to learn rich priors from input nighttime hazy images, thereby further improving the model's robustness and understanding for nighttime hazy scenes. For fine-tuning, we employ the pre-trained model from the synthetic domain to yield coarse haze-free images in an unsupervised fashion. Then, an efficient haze removal method called BCCR~\cite{meng2013efficient} is adopted to dehaze them for improving the visibility. Finally, the obtained pseudo ground truths are combined with real-world nighttime hazy images and fine-tuned in the synthetic domain to reduce the discrepancy between the synthetic and real domain. As shown in Fig.~\ref{comparisons1}(a)-(h), our NightHazeFormer produces a better dehazed result for a real-world nighttime hazy image. In addition, to bridge this gap between synthetic and real data, we have created a large-scale synthetic nighttime hazy image called~\textbf{UNREAL-NH}. Fig.~\ref{comparisons1}(i) depicts t-SNE map~\cite{van2008visualizing} of various synthetic datasets and real-world nighttime hazy image dataset (REAL-NH), which indicates that the simulated degradations of our UNREAL-NH are more realistic. Also, in Fig.~\ref{comparisons1}(j), the ``Fr$\acute{\mathrm{e}}$chet Inception Distance'' (FID) metric~\cite{heusel2017gans} that measures the distance between synthetic and real data at feature level quantitatively proves the superiority of our UNREAL-NH.

The main contributions are summarized as follows:
\begin{itemize}
\setlength{\topsep}{-3pt}
    \item We propose an end-to-end transformer-based network, called NightHazeFormer, for nighttime haze removal. By incorporating two powerful priors into the transformer decoder, our NightHazeFormer generates non-learnable prior queries that effectively guide our network to learn abundant prior features from input nighttime hazy images.
    \item A semi-supervised fine-tuning training paradigm is developed to improve the generalization ability. We combine the real-world nighttime hazy images with the generated pseudo ground truth labels, which are then fed into the synthetic domain to fine-tune the pre-trained model and enable it to learn the domain knowledge of real data.
    \item To compensate for the deficiencies of degradations simulation in existing datasets, we propose a large-scale synthetic nighttime hazy image dataset called UNREAL-NH. Our UNREAL-NH accounts for multiple types of degradations and addresses the limitations of existing datasets.
    \item Experimental results on several synthetic and real-world benchmarks demonstrate that our NightHazeFormer outperforms state-of-the-art nighttime dehazing methods both subjective visual comparisons and objective quality metrics.
\end{itemize}
\begin{figure*}[t]
   \vspace{0.5em}
   \setlength{\abovecaptionskip}{5pt}
   \setlength{\belowcaptionskip}{-8pt}
    \centering
    \includegraphics[width=0.90\linewidth]{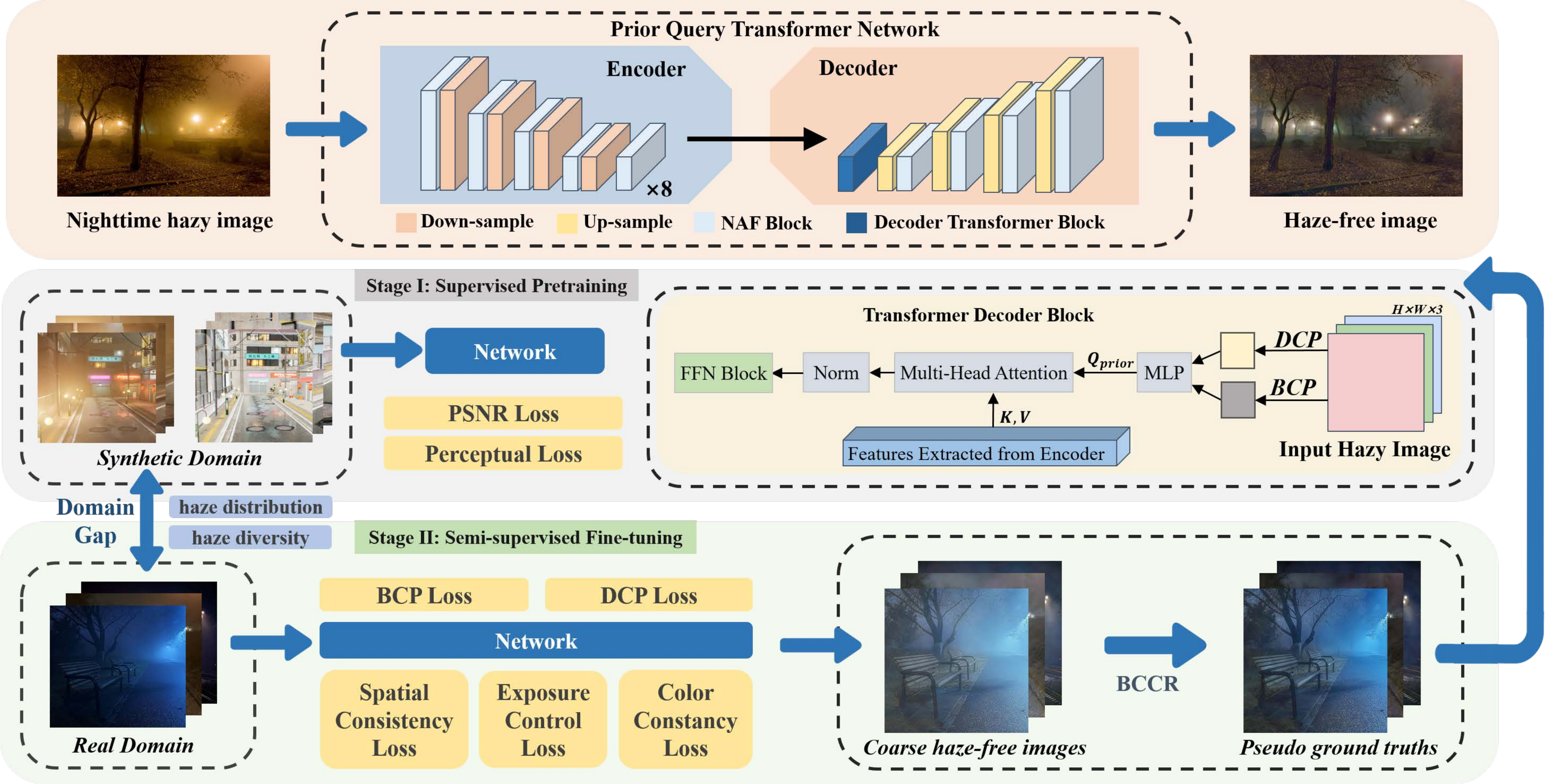}
    \caption{The architecture of our NightHazeFormer. Our approach comprises two stages: supervised pre-training and semi-supervised fine-tuning. For pre-training, we train our transformer network using paired synthetic images. The priors are incorporated into the transformer decoder to generate the prior queries, which guide the model to learn rich priors from input images. For fine-tuning, we devise a semi-supervised progressive refinement paradigm to improve the generalization ability. First, the unsupervised learning strategy allows our pre-trained model to fully leverage priors from real data and generate appropriate pseudo ground truths. Second, the generated pseudo-labels further enhance the model's generalization performance on the real domain in a cyclic supervised manner.}
    \label{fig2}
\end{figure*}
\section{Related Work}
\subsection{Daytime Dehazing Methods}
For daytime hazy scenarios, the imaging light sources are mainly dominated by the global airlight and the classic atmospheric scattering model~\cite{mccartney1976optics} is widely used to elucidate the degradation process of hazy images. To restore the haze-free image, earlier dehazing approaches usually make use of the priors or constraints (e.g. DCP~\cite{he2010single}, CAP~\cite{zhu2015fast}, BCCR~\cite{meng2013efficient}, color-lines~\cite{fattal2014dehazing}, haze-lines~\cite{berman2018single}, RLP~\cite{ju2021idrlp}, etc) to estimate the transmission and inverse the physical model to obtain the clear image. However, the presented priors may be invalid for diverse real-world scenes. Recently, with the rapid development of deep learning, numerous networks have been proposed to address computer vision tasks, such as image enhancement~\cite{HPEU,huang2022deep,ECLNet,Zhang_2023_CVPR,Huang_2023_CVPR,Huang_2022_CVPR,Huang_2018_ECCV_Workshops,Zheng2022unsupervised,jiang2023rsfdm,ye2022underwater}, pan-sharpening~\cite{
FourierPan,Zhou_2022_CVPR,Zhou2022effective,AdaptivePan2022MM,NormPan2022MM}, shadow removal~\cite{Zhu_2022_CVPR}, image desnowing~\cite{chen2022snowformer,chen2023msp,ye2022towards} and general image restoration~\cite{zhou2022deep,Zhang_2023_CVPR,Yang_2023_CVPR}.
To improve the haze removal performance, several efficient and effective dehazing networks have been developed to achieve the end-to-end mapping from a hazy image to a haze-free image, such as FFA-Net~\cite{qin2020ffa}, AECR-Net~\cite{wu2021contrastive}, PSD-Net~\cite{chen2021psd}, PMNet~\cite{ye2022perceiving}, PDD-GAN~\cite{chai2022pdd}, SFUDA~\cite{yu2022source}, etc. Although these aforementioned dehazing approaches perform well for daytime hazy scenarios, they are not effective to achieve quality improvement for nighttime hazy images. This is due to the fact that nighttime haze conditions usually contains multiple adverse effects and existing daytime dehazing methods cannot address these degradations.

\subsection{Nighttime Dehazing Methods}
To address the degradations presented in nighttime hazy scenes, Zhang~\emph{et al}.~\cite{zhang2014nighttime} construct a new imaging model to conduct nighttime dehazing. Considering the glow around artificial light sources, Li~\emph{et al}.~\cite{li2015nighttime} introduce a glow term into the atmospheric scattering model and perform the glow separation (GS). The maximum reflectance prior (MRP)~\cite{zhang2017fast} suited for nighttime hazy scenes has been developed to achieve fast restoration. Afterwards. Zhang~\emph{et al}.~\cite{zhang2020nighttime} devise an optimal-scale fusion-based dehazing (OSFD) method for nighttime hazy scenes. Liu~\emph{et al}. propose some variational-based decomposition models~\cite{liu2022nighttime,liu2023multi} to achieve structure dehazing and details enhancement. Unfortunately, these model-based algorithms only focus on partial degradations, which may result in unsatisfactory restoration results. On the other hand, deep learning based methods have been applied for nighttime image dehazing. Owing to requiring paired nighttime hazy images and clean images for training, Liao~\emph{et al}.~\cite{liao2018hdp} design two synthetic datasets (i.e. NightHaze and YellowHaze) by adding the haze into the collected nighttime images. Subsequently, several large-scale benchmarks are provided for nighttime dehazing, such as NHC~\cite{zhang2020nighttime}, NHM~\cite{zhang2020nighttime} and GTA5~\cite{yan2020nighttime}. Using these synthetic datasets, some networks, such as high-low frequency decomposition network~\cite{yan2020nighttime}, ND-Net~\cite{zhang2020nighttime} and GAPSF~\cite{jin2023enhancing}, are proposed to achieve nighttime dehazing. Recently, the universal dehazing networks, such as CFEN-ViT~\cite{zhao2021complementary} and RIDCP~\cite{wu2023ridcp}, are designed for both daytime and nighttime hazy scenes.

While learning-based nighttime image haze removal approaches have shown promising results for synthetic data, they usually struggle to generalize well to real-world nighttime hazy images. The main reasons for this are two-fold. First, there are significant inherent differences between previous generated synthetic dataset and real-world nighttime hazy images. Second, existing learning-based methods resort to training on synthetic datasets, which lacks the domain knowledge of the real data.

\section{Methodology}

\subsection{Framework Overview}\label{sec3.1}
In Fig.~\ref{fig2}, our framework consists of two stages: a supervised pre-training phase using prior query transformer network and a semi-supervised fine-tuning training phase using pseudo-labels.

\textbf{Supervised Pre-training.} For pre-training, we initially adopt the effective encoder-decoder transformer architecture with NAFBlocks~\cite{chen2022simple} as our backbone to learn the domain knowledge of nighttime hazy images. Due to the complex and multiple degradations presented in nighttime hazy images, we incorporate two powerful priors (i.e. DCP~\cite{he2010single} and BCP ~\cite{wang2013automatic}) into the transformer decoder to generate the non-learnable prior queries. Guided by prior information, the provided queries can effectively instruct the model to learn specific degradations from input nighttime hazy images, thereby enhancing the robustness of the model and understanding for nighttime hazy scenes. In this stage, the labeled synthetic data is solely used for supervised training to acquire a pre-trained model in the synthetic domain.


\textbf{Semi-supervised Fine-tuning.}
To improve the generalization ability of the pre-trained model, we propose a semi-supervised fine-tuning training paradigm based on the generated pseudo-labels. First, we perform training on unlabeled real data in an unsupervised manner to obtain the coarse haze-free images. Then, these results are further refined using an efficient daytime dehazing method named BCCR~\cite{meng2013efficient} to generate pseudo ground truths (GT). Finally, we combine the pseudo-GT labels with the corresponding real-world nighttime hazy images to form image pairs that are fed into the synthetic domain for fine-tuning the previous pre-trained model. This semi-supervised fine-tuning approach facilitates the acquisition of domain knowledge from real data, effectively improving the generalization performance to real-world nighttime hazy images.

\subsection{Prior Query Transformer Network}\label{sec3.2}

\textbf{Transformer Encoder.} Given an input nighttime hazy image $I$ with dimensions $H \times W \times 3$, we encode it into patches and feed forward them into the transformer encoder. For the design of the transformer encoder, we adopt NAFBlocks~\cite{chen2022simple} with excellent learning capability and high computational efficiency for feature extraction. The resolution of the feature maps is gradually reduced through down-sampling operations to extract multi-level features, thereby enabling the model to learn the hierarchical feature representation of the input image across scale. Notably, we observe that the feature maps with lower resolution are more effective in capturing global information, as each pixel represents a larger spatial region with richer information, such as shape, textures, and colors of the input image. Therefore, at the scale with the lowest resolution, we further employ eight NAFBlocks to learn the latent features and feed them to the transformer decoder.

\textbf{Transformer Decoder.} Since the nighttime hazy images usually contain complex and diverse degradations, it is necessary to introduce physical knowledge into the transformer decoder block to guide the model training. Previous studies~\cite{carion2020end,valanarasu2022transweather} have utilized the learnable queries to deal with detection and restoration tasks. Inspired by these methods, we incorporate two physical priors, namely DCP~\cite{he2010single} and BCP~\cite{wang2013automatic}, into the transformer decoder to generate the non-learnable queries. These non-learnable queries explicitly help the model to understand the degradations of nighttime hazy images and learn rich priors from input images, which can be calculated as follows:
\begin{equation}
    Q_{prior} = MLP(DCP(I)+BCP(I))
\end{equation}
where $MLP$ stands for the multi-layer perceptron. The usage of the DCP enables our network to focus on the hazy regions of degraded image, thus improving the dehazing ability. However, existing methods have demonstrated that the DCP may lead to darker results. To compensate for the deficiency of the DCP, we further incorporate the BCP to assist the model in learning the priors related to brightness features, thereby enhancing the contrast of the restored image. The combination of these two priors can improve the model's robustness and understanding for nighttime hazy images.

In this way, we utilize the non-learnable embedding of the prior knowledge as the queries ($Q$) of the multi-head attention and the latent features are used for keys ($K$) and values ($V$). The multi-head self-attention is calculated as follows:
\begin{equation}
    Attn(Q,K,V) = Softmax(\frac{QK^T}{\sqrt{d}})V
\end{equation}
where $d$ represents the dimension. The decoded features proficiently integrate the degradation features guided by physical priors, which provides sufficient guidance for the stripping of degradations in the subsequent process. These features are then passed through several up-sampling operations and fused with the corresponding features extracted from each stage in the transformer encoder to obtain the haze-free restoration results with dimensions $H \times W \times 3$. Similarly, we also adopt NAFBlocks in the transformer decoder to learn the high-level features of the input images.

\textbf{Loss Functions.} Our NightHazeFormer is optimized using two supervised loss functions. We first use the PSNR loss~\cite{chen2021hinet} as our basic restoration loss:
\begin{equation}
    \mathcal{L}_{psnr} = -PSNR(NightHazeFormer(I),J)
\end{equation}
where $J$ is the corresponding ground-truth of the input nighttime hazy image $I$.

Furthermore, we adopt the perceptual loss $\mathcal{L}_{per}$ to improve the visual quality of the restored results, which is calculated as follows:
\begin{equation}
    \mathcal{L}_{per} = \sum_{j=1}^2 \frac{1}{C_jH_jW_j} \Vert \phi_j(NightHazeFormer(I)) - \phi_j(J) \Vert_1
\end{equation}
where ${C_j}$, ${H_j}$ and ${W_j}$ respectively stand for the channel number, height and width of the feature map. $\phi_j$ represents the specified layer of VGG-19~\cite{simonyan2014very}.

Overall, the losses for supervised training can be expressed as:
\begin{equation}
    \mathcal{L}_{sl} = \mathcal{L}_{psnr} + \lambda_{per}\mathcal{L}_{per}
\end{equation}
where $\lambda_{per}$ is trade-off weight.

\subsection{Semi-supervised Fine-tuning Training}\label{sec3.3}
Owing to the inherent domain gap between synthetic and real data, existing nighttime haze removal networks solely trained on synthetic data suffer from limited generalization ability, resulting in unsatisfactory restoration results for real-world nighttime hazy images. To tackle this issue, we propose a semi-supervised fine-tuning training paradigm to help the pre-trained model narrow the discrepancy between synthetic and real domain. It consists of two phases: unsupervised learning using unlabeled real data and followed by supervised learning using pseudo-labels.

Specifically, the unlabeled real-world nighttime hazy images from our REAL-NH are first employed to train the model in an unsupervised manner. Through unsupervised learning, the model is able to better understand the degradations distribution and feature representation of the real data. However, due to the insufficient dehazing ability of the model trained on the synthetic domain, we further perform the supervised learning to fine-tune the pre-trained model based on the pseudo-labels. In order to generate pseudo ground truths (pseudo-GT), we make use of an efficient dehazing method called BCCR~\cite{meng2013efficient} to improve the quality of the dehazed results obtained from unsupervised training. By combining these generated pseudo-GT labels with real-world nighttime hazy images as paired data, we feed them into the synthetic domain to fine-tune the pre-trained model. This fine-tuning strategy enables the network to learn domain knowledge from real data, significantly enhancing the model's generalization performance.
\begin{figure}[!t]
   \vspace{0.5em}
   \setlength{\abovecaptionskip}{3pt}
   \setlength{\belowcaptionskip}{-13pt}
    \centering
    \includegraphics[width=0.95\linewidth]{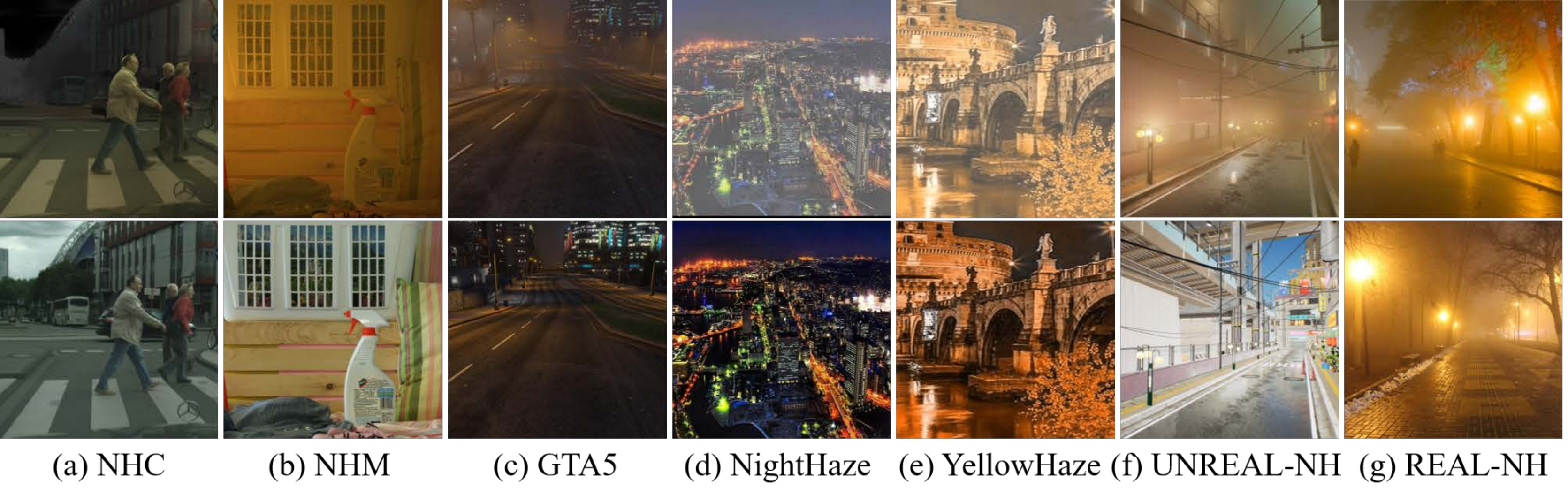}
    \caption{Visual comparison of various synthetic datasets. (a)-(f) represent the synthetic nighttime hazy patches and the corresponding clean patches extracted from NHC~\cite{zhang2020nighttime}, NHM~\cite{zhang2020nighttime}, GTA5~\cite{yan2020nighttime},  NightHaze~\cite{liao2018hdp}, YellowHaze~\cite{liao2018hdp}, and UNREAL-NH (Ours), respectively. (g) stands for real-world nighttime hazy patches extracted from REAL-NH.}
    \label{comparison_dataset}
\end{figure}

For the design of unsupervised losses, we follow the method proposed in~\cite{chen2021psd} and incorporate the classic DCP~\cite{he2010single} as the prior loss using an energy optimization function:
\begin{equation}
    \mathcal{L}_{dcp} = t^TLt+\lambda(t-\Tilde{t})^T(t-\Tilde{t})
\end{equation}
where $t$ and $\Tilde{t}$ respectively represent the transmission map estimated by the DCP and our model. $L$ is a Laplacian-like matrix. $\lambda$ is a hyper-parameter that controls the balance between the fidelity term and the penalty term. The DCP loss facilitates our network in acquiring the haze-related features from real-world nighttime hazy images.

However, $\mathcal{L}_{dcp}$ may lead to darker restoration results than expected. To overcome the drawbacks of $\mathcal{L}_{dcp}$, we further employ the effective BCP~\cite{wang2013automatic} as an additional prior loss to improve the brightness and contrast of the restoration result. The BCP loss is calculated as follows:
\begin{equation}
     \mathcal{L}_{bcp} = \Vert t-\Tilde{t} \Vert_1
\end{equation}
where $t$ and $\Tilde{t}$ respectively denote the transmission map estimated by the BCP and our model.

In addition, following the method~\cite{guo2020zero}, three types of losses, namely spatial consistency loss $\mathcal{L}_{spa}$ , exposure control loss $\mathcal{L}_{exp}$ and color constancy loss $\mathcal{L}_{col}$, are incorporated into the unsupervised loss functions to enhance the haze removal performance.

The unsupervised loss function ${{\mathcal L}_{ul}}$ can be expressed as follows:
\begin{equation}
\begin{array}{l}
{{\mathcal L}_{ul}} = {\lambda _{dcp}}{{\mathcal L}_{dcp}} + {\lambda _{bcp}}{{\mathcal L}_{bcp}}\\
\;\;\;\;\;\;\;\;\;\;\;\;\;\;\;\;\;\;\;\;\;\;\;\;\; + {\lambda _{spa}}{{\mathcal L}_{spa}} + {\lambda _{exp}}{{\mathcal L}_{exp}} + {\lambda _{col}}{{\mathcal L}_{col}}
\end{array}
\end{equation}
where $\lambda_{dcp}$, $\lambda_{bcp}$, $\lambda_{spa}$, $\lambda_{exp}$ and $\lambda_{col}$ are trade-off weights.

\begin{figure}[!t]
   \vspace{0.5em}
   \setlength{\abovecaptionskip}{3pt}
   \setlength{\belowcaptionskip}{-17pt}
    \centering
    \centering
    \includegraphics[width=0.95\linewidth]{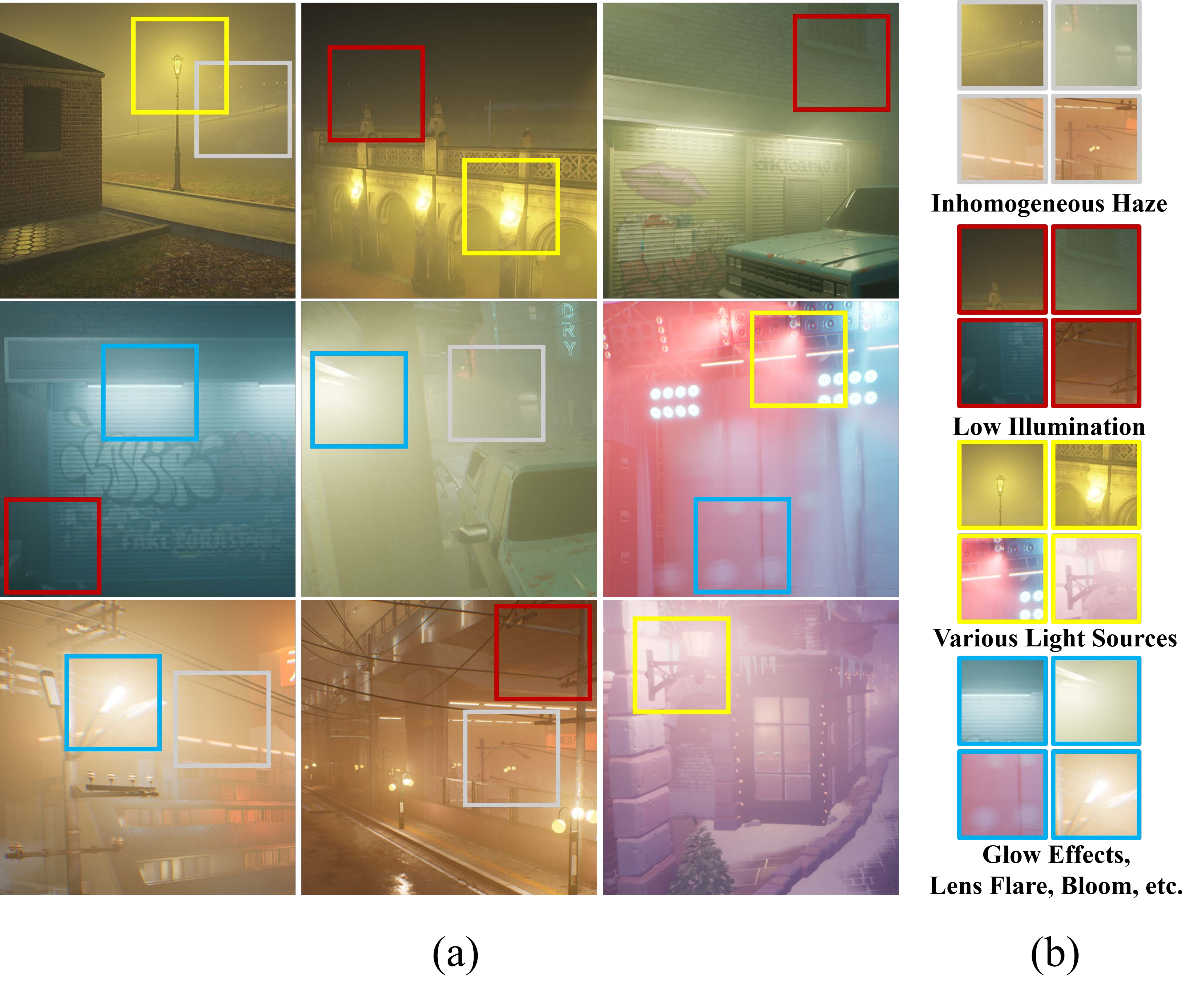}
    \caption{(a) Examples of our UNREAL-NH. (b) Examples of multiple degradations extracted from (a).}
    \label{degradations_dataset}
\end{figure}
\begin{figure*}[htb]
    \vspace{0.5em}
    \setlength{\abovecaptionskip}{3pt}
    \setlength{\belowcaptionskip}{-10pt}
    \centering
    \includegraphics[width=0.95\linewidth]{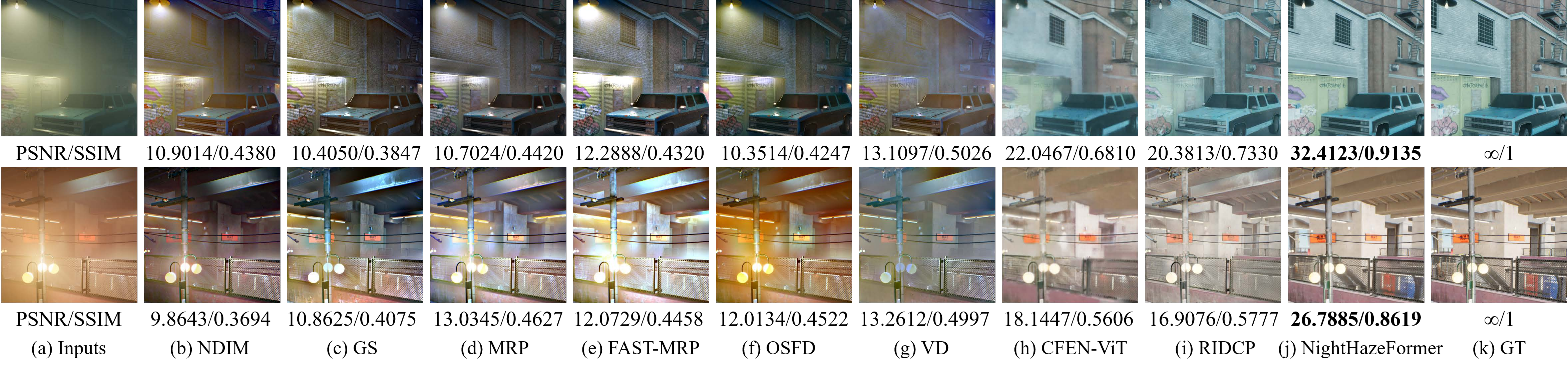}
    \caption{Visual comparisons on synthetic nighttime hazy images images from our UNREAL-NH.}
    \label{qualitative_UNREAL_NH}
\end{figure*}
\begin{figure*}[htb]
    \vspace{0.5em}
    \setlength{\abovecaptionskip}{3pt}
    \setlength{\belowcaptionskip}{-10pt}
    \centering
    \includegraphics[width=0.95\linewidth]{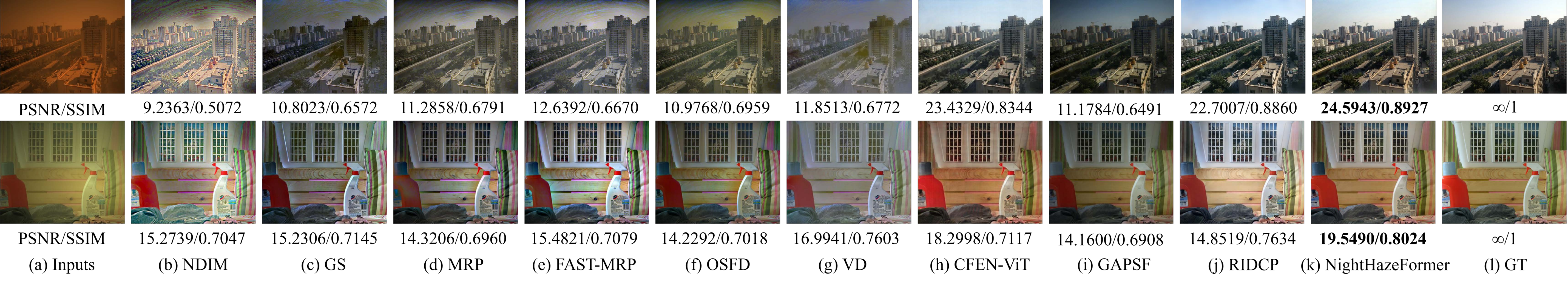}
    \caption{Visual comparisons on synthetic nighttime hazy images from NHR (first row) and NHM (last row).}
    \label{qualitative_NHR_NHM}
\end{figure*}

\section{Experiments}
\subsection{Dataset Generation}
Existing synthetic datasets for nighttime image dehazing, including NHC~\cite{zhang2020nighttime}, NHM~\cite{zhang2020nighttime}, GTA5~\cite{yan2020nighttime}, NightHaze~\cite{liao2018hdp} and YellowHaze~\cite{liao2018hdp}, only incorporate a limited range of degradations and fail to adequately simulate various point sources, the surrounding glow, hidden noise, blurring effects, etc. As a result, these datasets show significant disparities from real-world nighttime hazy scenes, as shown in Fig.~\ref{comparison_dataset}. To overcome the limitations of the above datasets, we utilized \textbf{UNREAL} Engine 4.27~\cite{UE} to construct a large-scale paired synthetic dataset of \textbf{N}ighttime \textbf{H}azy images, called \textbf{UNREAL-NH}. Fig.~\ref{degradations_dataset} illustrates some examples of synthetic nighttime hazy images from our UNREAL-NH dataset, accompanied by multiple critical degradation effects we have considered. In addition, we employ FID metric~\cite{heusel2017gans} to objectively measure the difference between the constructed synthetic dataset and real-world dataset, as illustrated in Fig.~\ref{comparisons1}(j). A smaller FID value indicates a closer resemblance to the real domain. Compared to previous datasets, our proposed UNREAL-NH contains most of the common degradation factors and significantly simulates real-world nighttime hazy scenes.

To be specific, we first adopt various fog effects and add multiple point light sources with different colors in the Unreal Engine to create 1260 pairs of synthetic nighttime hazy images and the corresponding clean images with a resolution of $2008 \times 1129$. Then, several post-processing techniques, such as motion blur, lens flare and bloom, are applied to the generated synthetic nighttime hazy images to make them more realistic. Finally, to facilitate training, the random overlap cropping strategies are adopted to generate 10080 pairs with a resolution of $480 \times 480$.

Moreover, we also have collected 250 \textbf{REAL}-world \textbf{N}ighttime \textbf{H}azy images called \textbf{REAL-NH}, in which 150 images are from NHRW~\cite{zhang2020nighttime} and other 100 images are collected from the Internet.

\subsection{Experimental Settings}
\textbf{Datasets.} For pre-training, our UNREAL-NH dataset is used for supervised learning, which is split into a training set with 8064 image pairs, a validation set with 1008 image pairs and a test set with 1008 image pairs. For fine-tuning, we select 200 real-world nighttime hazy images from our REAL-NH dataset for semi-supervised learning, while the remaining 50 images are used for testing. In addition, we also train our NightHazeFormer on 8073 image pairs from NHR dataset following~\cite{zhang2020nighttime}. Subsequently, we conduct the quantitative evaluation on a test set of 897 images from NHR dataset and the complete NHM datesets (including 350 images)~\cite{zhang2020nighttime} to demonstrate the superiority of our proposed NightHazeFormer.

\textbf{Evaluation Metrics.} We utilize Peak Signal-to-Noise Ratio (PSNR) and Structural Similarity Index (SSIM)~\cite{wang2004image} to assess the dehazing results on the synthetic datasets (UNREAL-NH, NHR and NHM). Moreover, two non-reference image quality assessment metrics, namely NIQE~\cite{mittal2012making} and MUSIQ~\cite{ke2021musiq}, are adopted to evaluate the dehazed results on REAL-NH test dataset for quantitative comparisons. We choose MUSIQ model trained on one aesthetics quality dataset (AVA~\cite{murray2012ava}). The code of non-reference metrics is available on the github\footnote{https://github.com/chaofengc/Awesome-Image-Quality-Assessment}.

\textbf{Implementation Details.} Our framework is implemented using PyTorch~\cite{paszke2019pytorch} and trained on an NVIDIA RTX 3090 GPU (24GB) with a batch size of 16. To augment the training data in UNREAL-NH, we randomly crop the images into patches with a size of $256 \times 256$ and apply random rotations of 90, 180, 270 degrees, as well as the horizontal flip. In the supervised training, we adopt Adam optimizer with an initial learning rate of $2\times 10^{-4}$, $\beta_{1}= 0.9$ and $\beta_{2}= 0.999$. For unsupervised training, the initial learning rate is set to $5\times 10^{-5}$. In the training process, we also employ the Cyclic Learning Rate (CyclicLR) with a maximum learning rate of 1.2 times the initial learning rate. The trade-off weights $\lambda_{per}$, $\lambda_{dcp}$, $\lambda_{bcp}$, $\lambda_{spa}$, $\lambda_{exp}$ and $\lambda_{col}$ are set to $0.2$, $10^{-4}$, $10^{-4}$, $5$, $10^{-3}$, $0.2$, respectively.
\begin{table*}[!t]
\vspace{0.0em}
\setlength{\abovecaptionskip}{5pt}
    \caption{Quantitative comparisons of state-of-the-art methods on the UNREAL-NH, REAL-NH, NHR and NHM datasets.}\label{hazyresults}
    \centering
    \resizebox{16cm}{!}{
    \renewcommand\arraystretch{1.1}
    \begin{tabular}{c||c|c|cc|cc|cc|cc}
    \toprule[0.4pt]
    \gr
    & & & \multicolumn{2}{c|}{ \textbf{UNREAL-NH} } &\multicolumn{2}{c|}{ \textbf{REAL-NH} } & \multicolumn{2}{c|}{ \textbf{NHR}~\cite{zhang2020nighttime} } & \multicolumn{2}{c}{\textbf{NHM}~\cite{zhang2020nighttime}} \\ \cline{4-11}
    \gr
    \multirow{-2}{*}{\textbf{Type}} & \multirow{-2}{*}{\textbf{Method}} & \multirow{-2}{*}{\textbf{Venue}} & PSNR $\uparrow$ & SSIM $\uparrow$ & NIQE $\downarrow$ & MUSIQ-AVA $\uparrow$  & PSNR $\uparrow$ & SSIM $\uparrow$ & PSNR $\uparrow$ & SSIM $\uparrow$  \\ \hline
     \midrule[0.4pt]
     & NDIM~\cite{zhang2014nighttime}& \textit{ICIP'2014} & $9.3953$ & $0.4143$ & $4.0133$ & $4.9710$ & $11.5749$ & $0.5971$ & $12.6878$ &  $0.6171$ \\
     & GS~\cite{li2015nighttime} & \textit{ICCV'2015} & $9.1722$ & $0.3926$ & $3.8445$ & $4.9599$& $13.1634$ & $0.6266$ & $11.8528$ & $0.6148$  \\
    Model-based methods & MRP~\cite{zhang2017fast}& \textit{CVPR'2017} &$9.9146$ & $0.4391$ & $4.0657$ & $5.0461$ & $12.0909$ & $0.6955$ & $13.1088$ & $0.6545$  \\
    & FAST-MRP~\cite{zhang2017fast} & \textit{CVPR'2017}& $10.7856$ &$0.4488$ & $4.0831$ & $4.9309$ & $13.5419$ & $0.6837$ & $13.3081$ & $0.6491$  \\
    & OSFD~\cite{zhang2020nighttime} &  \textit{MM'2020} & $9.0816$ & $0.4227$ & $4.0853$ & $5.0961$ & $11.8953$ & $0.6923$ & $13.2819$ &$0.6667$ \\
     &VD~\cite{liu2022nighttime} & \textit{CVPRW'2022} & $11.2984$ &$0.5171$ & $3.9231$ & $5.1223$ & $13.2572$& $0.7267$ & $13.7675$ & $0.6865$ \\
    \hline
     & CFEN-ViT~\cite{zhao2021complementary} & \textit{Arxiv'2021} & $19.1704$ & $0.6254$ & $3.8466$ & $4.8755$ & $21.5380$ & $0.8150$ & $16.9481$ & $0.7224$  \\
  Learning-based methods & GAPSF~\cite{jin2023enhancing} & \textit{Arxiv'2023} & - & - & 4.1208 & 4.6300 & 13.1660 & 0.6945 & 13.4808 & 0.6653  \\
      & RIDCP~\cite{wu2023ridcp} & \textit{CVPR'2023} & $17.6887$ & $0.6456$ & $3.7856$ & $5.2590$ & $21.8923$ & $0.8777$ & $14.8597$ & $0.7516$ \\
    \hline \hline
   Learning-based method & NightHazeFormer (Ours) & \textit{MM'2023} & $\mathbf{27.9277}$ & $\mathbf{0.8669}$ & $\mathbf{3.6811}$ & $\mathbf{5.2635}$ & $\mathbf{23.6649}$ & $\mathbf{0.9031}$ & $\mathbf{18.5400}$ & $\mathbf{0.7842}$\\

    \bottomrule[0.4pt]
    \end{tabular}}
    \label{table:comparisons}
\end{table*}
\begin{figure*}[!t]
    \vspace{0.5em}
    \setlength{\abovecaptionskip}{3pt}
    \setlength{\belowcaptionskip}{-8pt}
    \centering
    \includegraphics[width=0.95\linewidth]{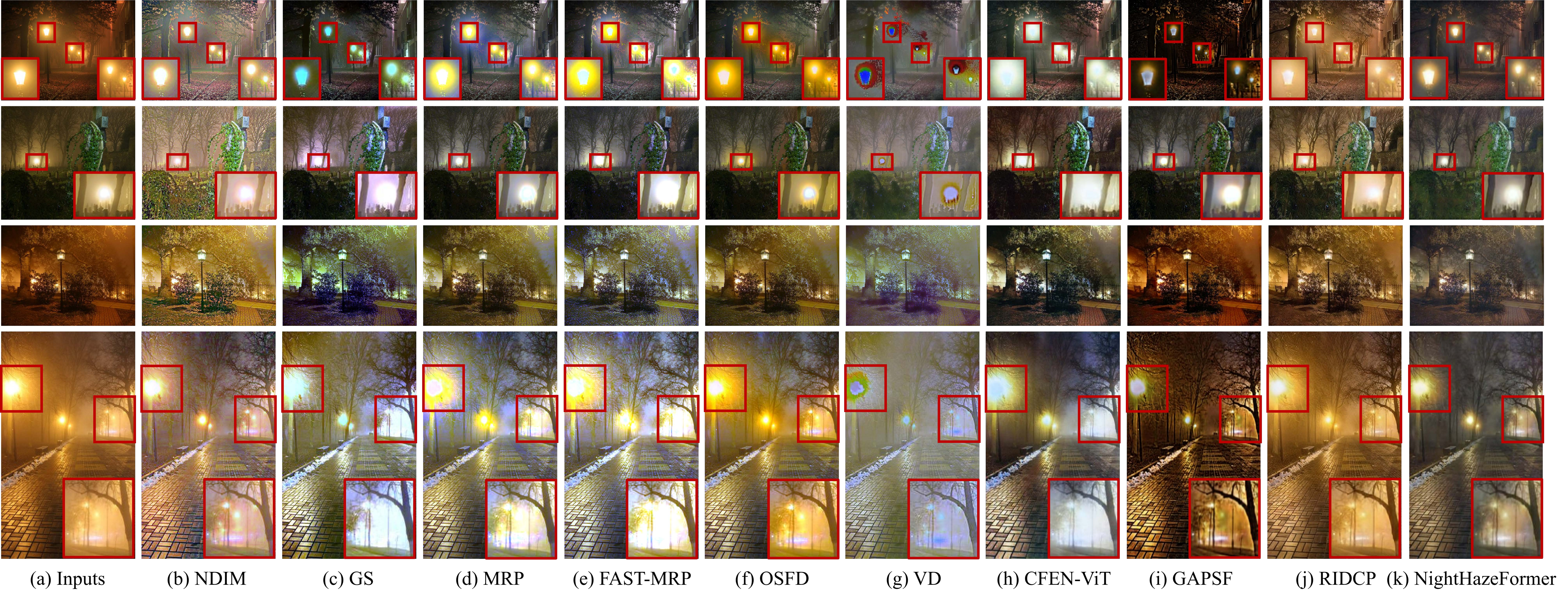}
    \caption{Visual comparisons on real-world nighttime hazy images from REAL-NH dataset.}
    \label{comparisons_real}
\end{figure*}
\subsection{Comparisons with State-of-the-art Methods}
To verify the effectiveness and generalization ability of our NightHazeFormer, we conduct comparisons with several state-of-the-art specific nighttime dehazing methods, including model-based methods (i.e. NDIM~\cite{zhang2014nighttime}, GS~\cite{li2015nighttime}, MRP~\cite{zhang2017fast}, FAST-MRP~\cite{zhang2017fast}, OSFD~\cite{zhang2020nighttime} and VD~\cite{liu2022nighttime}) and learning-based methods (i.e. CFEN-ViT~\cite{zhao2021complementary}, GAPSF~\cite{jin2023enhancing} and RIDCP~\cite{wu2023ridcp}) on synthetic and real-world datasets. The dehazed results of GAPSF on NHM, NHR and REAL-NH test datasets are provided by the authors. For CFEN-ViT and RIDCP, we retrain their released code on the training set of synthetic datasets (UNREAL-NH and NHR) and then employ the retrained models on the corresponding testing sets to ensure fair comparisons. Since CFEN-ViT and RIDCP have the domain adaptation capability for nighttime dehazing, their released pre-trained models are adopted on real-world dataset (REAL-NH) for comparisons.

\textbf{Visual Comparisons on Synthetic and Real-world Images.} The visual comparisons on synthetic nighttime hazy images from UNREAL-NH, NHR and NHM datasets are illustrated in Fig.~\ref{qualitative_UNREAL_NH} and Fig.~\ref{qualitative_NHR_NHM}. It is clearly observed that the results of NDIM, GS, MRP, FAST-MRP, OSFD and VD not only fail to overcome multiple degradations but also suffer from halo artifacts. GAPSF tends to yield dehazed results that appear darker with fewer details. Although CFEN-ViT and RIDCP effectively remove nighttime haze, they still encounter challengings in restoring fine details. In contrast, our NightHazeFormer shows promising performance in nighttime haze removal and details restoration. Furthermore, our results achieve the highest PSNR and SSIM values for test images.

In addition, we also evaluate the visual effects on real-world nighttime hazy image from REAL-NH dataset in Fig.~\ref{comparisons_real}. From the visual comparisons, we find that NDIM, GS, MRP, FAST-MRP, OSFD and VD fail to restore the details and remove the glow around the artificial light sources. The learning-based methods, such as CFEN-ViT and RIDCP, also struggle with handling glow effects due to their insufficient generalization performance. GAPSF is effective in mitigating glow around lights, but this method tends to produce results with color shifts and darkness. Compared to the aforementioned methods, our NightHazeFormer simultaneously achieves haze removal, glow suppression, details restoration and color correction which demonstrates superior generalization ability.

\textbf{Quantitative Comparisons on Synthetic and Real-World Datasets.}
Table~\ref{hazyresults} reports the quantitative comparisons on the testing set of three synthetic datasets (UNREAL-NH, NHR and NHM ) and a real-world dataset (REAL-NH). For synthetic datasets (UNREAL-NH, NHR and NHM), the clear images instead of low-light images are considered as the reference images to calculate PSNR and SSIM. In addition, the results obtained by different methods have variations in resolution, which may lead to discrepancies when calculating non-reference image quality metrics. To ensure fair comparisons, the dehazed results of all methods on the REAL-NH test dataset are resized to $600 \times 400$ for objective evaluation. As depicted in Table~\ref{hazyresults}, our NightHazeFormer outperforms all the compared methods in terms of PSNR and SSIM values on the UNREAL-NH, NHR, NHM datasets. Moreover, it achieves the best MUSIQ-AVA and NIQE scores on the REAL-NH test dataset, verifying its outstanding generalization performance.

\subsection{Ablation Studies}
In order to demonstrate the effectiveness of the design in our proposed NightHazeFormer, a series of ablation studies are performed.

\begin{figure}[!t]
    \vspace{0.5em}
    \setlength{\abovecaptionskip}{2pt}
    \setlength{\belowcaptionskip}{-13pt}
    \centering
    \includegraphics[width=0.95\linewidth]{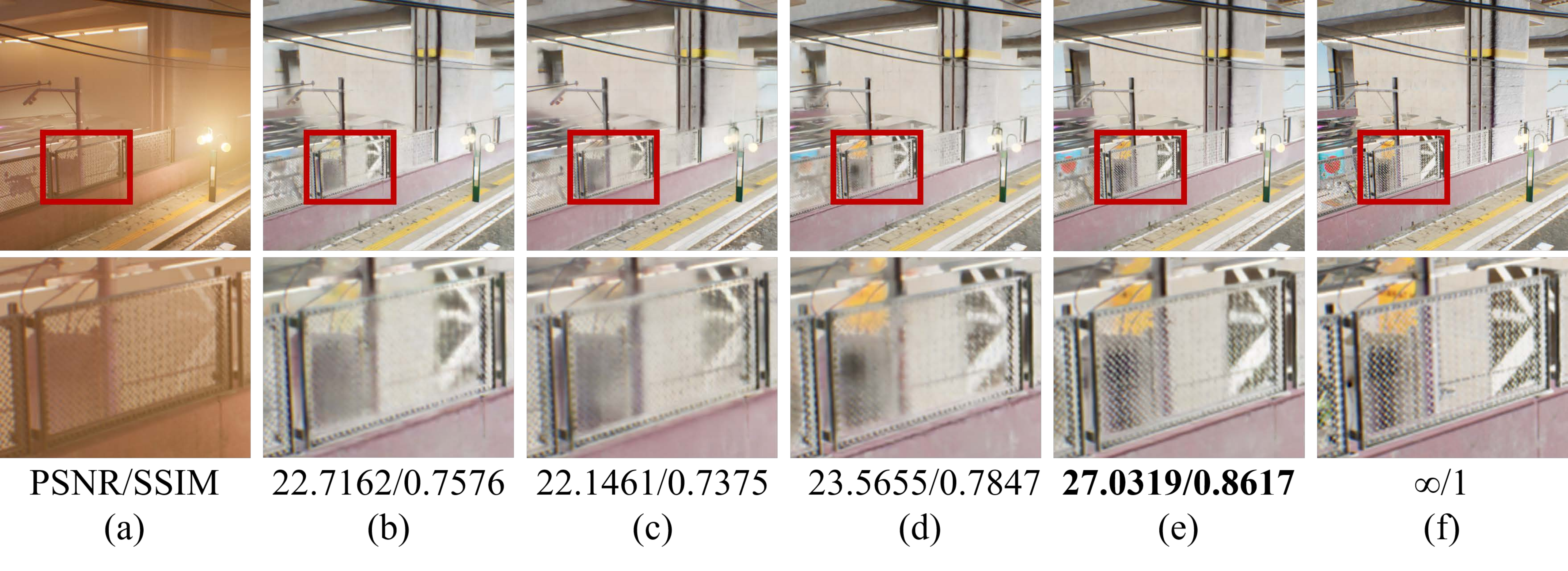}
    \caption{Visual results of ablation study on NAFBlock module. (a) Input. (b) ResBlock\cite{he2016deep}. (c) ViTBlock~\cite{dosovitskiy2020image}. (d) without latent features extraction. (e) NAFBlock (Ours). (f) GT.}
    \label{ablation_blocks}
\end{figure}

\textbf{Effectiveness of NAFBlock Module.}
The usage of NAFBlock contributes to features extraction with high computational efficiency. To prove the contribution of the NAFBlock module, two well-known modules, ResBlock~\cite{he2016deep} and ViTBlock~\cite{dosovitskiy2020image}, are used to replace the NAFBlock in the transformer decoder. In addition, we also employ eight NAFBlocks in the transformer encoder to extract the latent features from the image with the lowest resolution. The latent features modeling helps our model capture richer global information. To demonstrate this point, we conduct ablation study by removing eight NAFBlocks from the transformer encoder. The quantitative comparisons presented in Table~\ref{Table:blocks} and visual results depicted in Fig.~\ref{ablation_blocks} demonstrate the significance of latent features modeling and the effectiveness of the NAFBlock module.

\begin{figure}[!t]
    \vspace{0.5em}
    \setlength{\abovecaptionskip}{2pt}
    \setlength{\belowcaptionskip}{-13pt}
    \centering
    \includegraphics[width=0.95\linewidth]{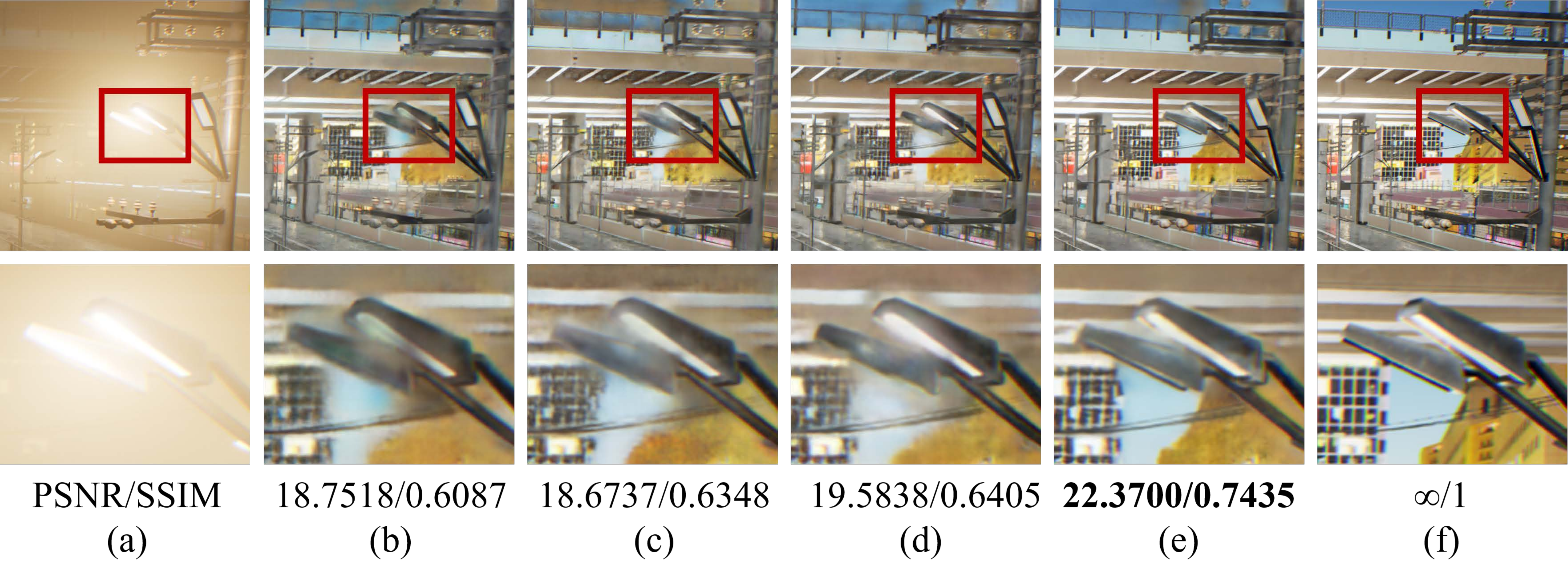}
    \caption{Visual results of ablation study on prior queries. (a) Input. (b) without priors. (c) with $Q_{DCP}$. (d) with $Q_{BCP}$. (e) with $Q_{Prior}$ (Ours). (f) GT.}
    \label{ablation_priors}
\end{figure}

\textbf{Effectiveness of Prior Queries.}
The non-learnable prior queries $Q_{Prior}$ in the transformer decoder generated by two powerful priors guide the model to extract specific degradations. To verify the effectiveness of two priors, we consider the basic encoder-decoder architecture without any priors as our baseline and then introduce the DCP and BCP separately into the baseline to generate the prior queries $Q_{DCP}$ and $Q_{BCP}$, respectively. Table~\ref{Table:prior} illustrates the quantitative comparisons of various settings on the UNREAL-NH test dataset, which indicates that the combination of two priors can effectively improve the haze removal performance. Furthermore, the visual results in Fig.~\ref{ablation_priors} also demonstrate the effectiveness of the generated non-learnable prior queries.

\textbf{Effectiveness of Semi-supervised Fine-tuning.} Our proposed framework mainly consists of two stages: supervised learning and semi-supervised fine-tuning. To verify the contribution of the semi-supervised fine-tuning stage, an ablation study is performed involving three settings: 1) Setting i: conducting only supervised learning, 2) Setting ii: conducting both supervised and unsupervised learning, 3) Setting iii: conducting supervised and unsupervised learning, followed by fine-tuning based on the generated pseudo-labels. In Fig.~\ref{semi_supervised}(b)-(c), the dehazing model trained with both supervised and unsupervised learning demonstrates superior generalization ability for real-world nighttime hazy scenes. Compared Fig.~\ref{semi_supervised}(c) with Fig.~\ref{semi_supervised}(d), our full model with fine-tuning stage contributes to dehazing performance improvement.
\begin{table}[!t]
\vspace{2.5em}
\setlength{\abovecaptionskip}{5pt}
\setlength{\belowcaptionskip}{-16pt}
\centering
\caption{Ablation study on NAFblock module.}
\begin{tabular}{c|cc}
\hline
\gr
 & \multicolumn{2}{c}{UNREAL-NH}                                      \\ \cline{2-3}

\gr  \multirow{-2}{*}{Setting}                        & PSNR$\uparrow$ & SSIM$\uparrow$  \\ \hline
ResBlock                 & $23.9723$     & $0.7786$                  \\
ViTBlock                 & $23.5148$     & $0.7668$                 \\
w/o latent features      & $24.8271$     & $0.8038$              \\
NAFBlock (Ours)          & $\textbf{27.9277}$ & $\textbf{0.8669}$  \\ \hline
\end{tabular}
\label{Table:blocks}
\end{table}
\begin{table}
\setlength\tabcolsep{2.5pt}
\vspace{0.5em}
\setlength{\abovecaptionskip}{3pt}
\setlength{\belowcaptionskip}{-10pt}
\centering
\caption{Ablation study on prior queries.}
\begin{tabular}{c|cc}
\hline
\gr
 & \multicolumn{2}{c}{UNREAL-NH}       \\ \cline{2-3}

\gr  \multirow{-2}{*}{Setting}                        & PSNR$\uparrow$ & SSIM$\uparrow$  \\ \hline
  Baseline              &  $24.8434$    &  $0.7993$            \\
  Baseline + $Q_{DCP}$ &  $25.2372$    &  $0.8139$           \\
Baseline + $Q_{BCP}$ &  $25.2761$    &  $0.8149$          \\
Baseline + $Q_{Prior}$ (Ours) &  $\textbf{27.9277}$ & $\textbf{0.8669}$    \\ \hline
\end{tabular}
\label{Table:prior}
\end{table}

\textbf{Effectiveness of Unsupervised Losses.} To demonstrate the contribution of our unsupervised loss committee, we conduct the ablation experiments as follows: (a) without spatial consistency loss $\mathcal{L}_{spa}$, (b) without exposure control loss $\mathcal{L}_{exp}$, (c) without color constancy loss $\mathcal{L}_{col}$, (d) without DCP loss $\mathcal{L}_{dcp}$, (e) without BCP loss $\mathcal{L}_{bcp}$, and (f) with all losses. As viewed in Fig.~\ref{unsupervised_losses}(a), the loss committee without $\mathcal{L}_{spa}$ results in poor contrast and details restoration. In Fig.~\ref{unsupervised_losses}(b) and Fig.~\ref{unsupervised_losses}(e), without $\mathcal{L}_{exp}$ or $\mathcal{L}_{bcp}$, the dehazed result appears to be darker than expected. Fig.~\ref{unsupervised_losses}(c) shows that the absence of $\mathcal{L}_{col}$ leads to the color shifts. From Fig.~\ref{unsupervised_losses}(d), the lack of $\mathcal{L}_{dcp}$ makes the results blurred with residue hazes. We can observe that our method with all unsupervised losses generates visually satisfactory haze removal result.
\begin{figure}[]
    \vspace{0.5em}
    \setlength{\abovecaptionskip}{1pt}
    \setlength{\belowcaptionskip}{-16pt}
    \centering
    \includegraphics[width=0.95\linewidth]{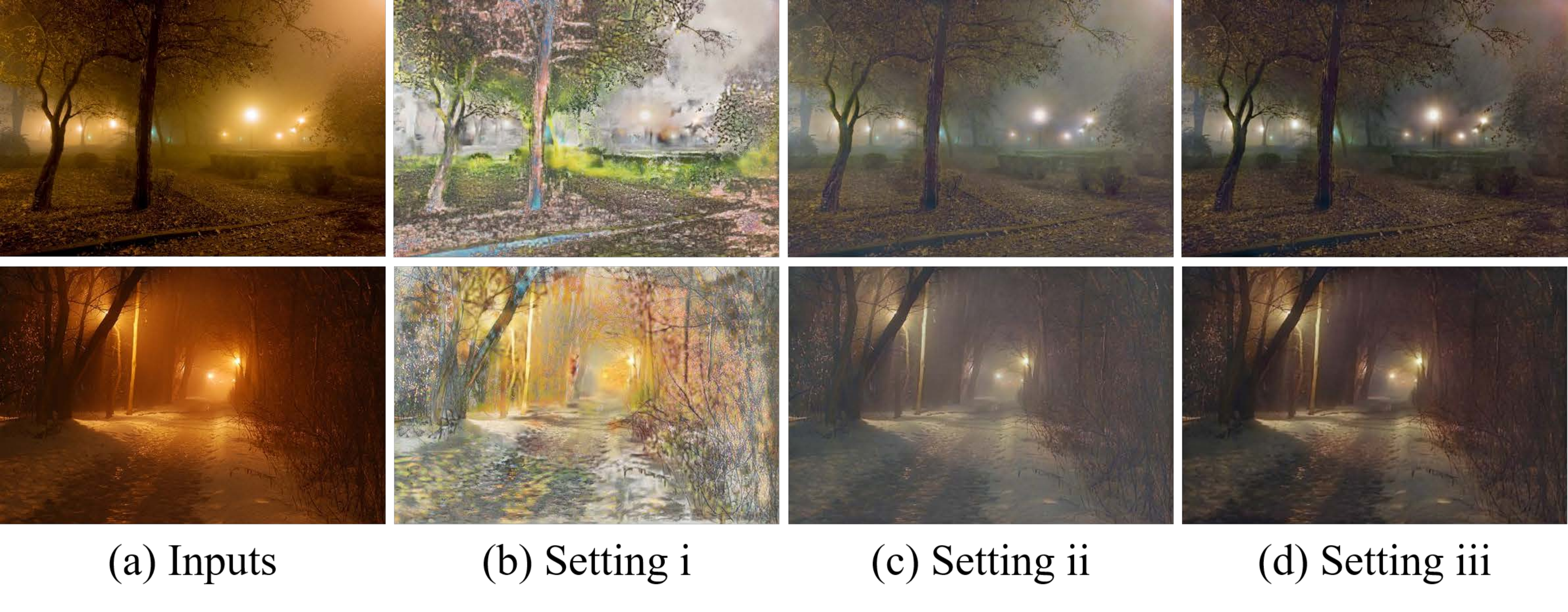}
    \caption{Visual results of ablation study on the semi-supervised fine-tuning.}
    \label{semi_supervised}
\end{figure}
\begin{figure}[]
    \vspace{1.0em}
    \setlength{\abovecaptionskip}{5pt}
    \setlength{\belowcaptionskip}{-10pt}
    \centering
    \includegraphics[width=0.95\linewidth]{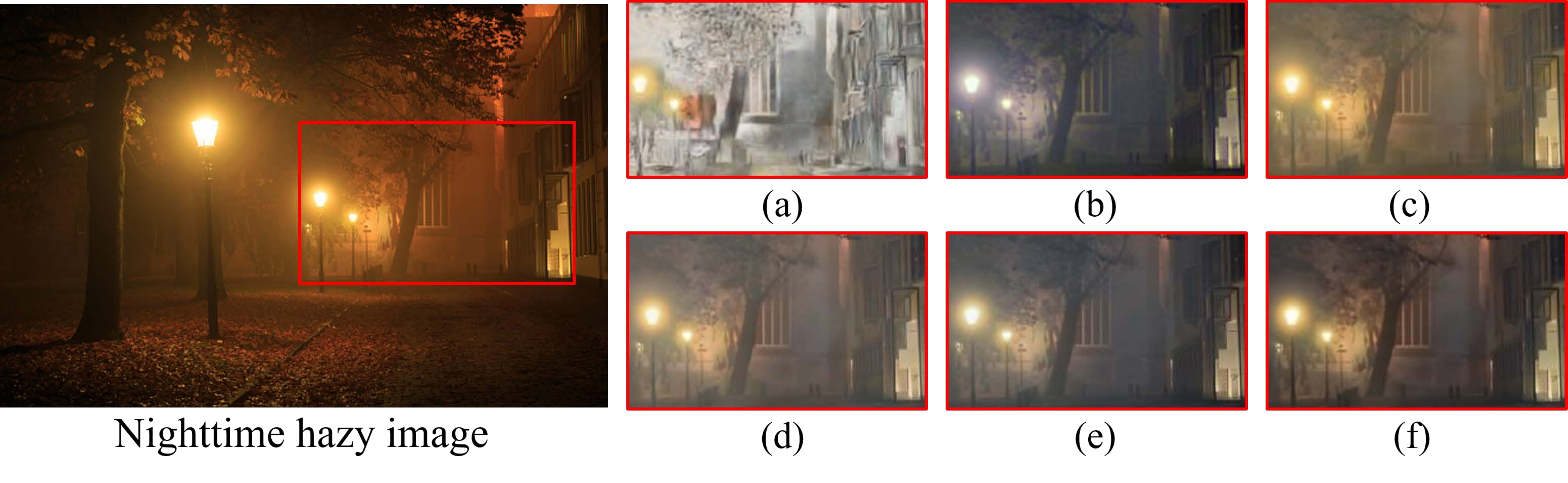}
    \caption{Visual results of ablation study on the unsupervised losses. (a) without $\mathcal{L}_{spa}$. (b) without $\mathcal{L}_{exp}$. (c) without $\mathcal{L}_{col}$. (d) without $\mathcal{L}_{dcp}$. (e) without $\mathcal{L}_{bcp}$. (f) Our method. }
    \label{unsupervised_losses}
\end{figure}

\section{Conclusion}
In this paper, we have proposed NightHazeFormer, a two-stage transformer-based network for nighttime haze removal. Previous model-based and learning-based algorithms have the limitations to address multiple degradations presented in real-world nighttime haze scenarios. To circumvent the above problem, we integrate two well-known priors into the transformer decoder to provide the prior queries for specific degradations extraction. Then, we develop a semi-supervised fine-tuning training paradigm to improve the generalization performance. Specifically, the generated pseudo ground truths and the real-world nighttime hazy images are paired together and then fed into the synthetic domain to fine-tune the pre-trained model. This procedure significantly contributes to acquiring the domain knowledge from real data. Besides, to address the issue of unrealistic degradations simulation in existing synthetic datasets for nighttime haze removal, we construct a large-scale nighttime hazy image dataset called UNREAL-NH for training data. Extensive experiments demonstrate that our proposed NightHazeFormer achieves superior haze removal performance and generalization ability over all other state-of-the-art dehazing methods in terms of qualitative and quantitative comparisons.

\section{ACKNOWLEDGMENTS}
This work was supported in part by the Open Research Fund Program of Data Recovery Key Laboratory of Sichuan Province (Grant No. DRN2306) and the Natural Science Foundation of Fujian Province (Grant No. 2021J01867).

\bibliographystyle{ACM-Reference-Format}
\balance
\bibliography{refs}


\begin{thebibliography}{75}


\ifx \showCODEN    \undefined \def \showCODEN     #1{\unskip}     \fi
\ifx \showDOI      \undefined \def \showDOI       #1{#1}\fi
\ifx \showISBNx    \undefined \def \showISBNx     #1{\unskip}     \fi
\ifx \showISBNxiii \undefined \def \showISBNxiii  #1{\unskip}     \fi
\ifx \showISSN     \undefined \def \showISSN      #1{\unskip}     \fi
\ifx \showLCCN     \undefined \def \showLCCN      #1{\unskip}     \fi
\ifx \shownote     \undefined \def \shownote      #1{#1}          \fi
\ifx \showarticletitle \undefined \def \showarticletitle #1{#1}   \fi
\ifx \showURL      \undefined \def \showURL       {\relax}        \fi
\providecommand\bibfield[2]{#2}
\providecommand\bibinfo[2]{#2}
\providecommand\natexlab[1]{#1}
\providecommand\showeprint[2][]{arXiv:#2}

\bibitem[Berman et~al\mbox{.}(2020)]%
        {berman2018single}
\bibfield{author}{\bibinfo{person}{Dana Berman}, \bibinfo{person}{Tali
  Treibitz}, {and} \bibinfo{person}{Shai Avidan}.}
  \bibinfo{year}{2020}\natexlab{}.
\newblock \showarticletitle{Single image dehazing using haze-lines}.
\newblock \bibinfo{journal}{\emph{IEEE Transactions on Pattern Analysis and
  Machine Intelligence}} \bibinfo{volume}{42}, \bibinfo{number}{3}
  (\bibinfo{year}{2020}), \bibinfo{pages}{720--734}.
\newblock


\bibitem[Bui and Kim(2018)]%
        {bui2018single}
\bibfield{author}{\bibinfo{person}{Trung~Minh Bui} {and} \bibinfo{person}{Wonha
  Kim}.} \bibinfo{year}{2018}\natexlab{}.
\newblock \showarticletitle{Single image dehazing using color ellipsoid prior}.
\newblock \bibinfo{journal}{\emph{IEEE Transactions on Image Processing}}
  \bibinfo{volume}{27}, \bibinfo{number}{2} (\bibinfo{year}{2018}),
  \bibinfo{pages}{999--1009}.
\newblock


\bibitem[Cai et~al\mbox{.}(2016)]%
        {cai}
\bibfield{author}{\bibinfo{person}{Bolun Cai}, \bibinfo{person}{Xiangmin Xu},
  \bibinfo{person}{Kui Jia}, \bibinfo{person}{Chunmei Qing}, {and}
  \bibinfo{person}{Dacheng Tao}.} \bibinfo{year}{2016}\natexlab{}.
\newblock \showarticletitle{DehazeNet: An end-to-end system for single image
  haze removal}.
\newblock \bibinfo{journal}{\emph{IEEE Transactions on Image Processing}}
  \bibinfo{volume}{25}, \bibinfo{number}{11} (\bibinfo{year}{2016}),
  \bibinfo{pages}{5187--5198}.
\newblock


\bibitem[Carion et~al\mbox{.}(2020)]%
        {carion2020end}
\bibfield{author}{\bibinfo{person}{Nicolas Carion}, \bibinfo{person}{Francisco
  Massa}, \bibinfo{person}{Gabriel Synnaeve}, \bibinfo{person}{Nicolas
  Usunier}, \bibinfo{person}{Alexander Kirillov}, {and} \bibinfo{person}{Sergey
  Zagoruyko}.} \bibinfo{year}{2020}\natexlab{}.
\newblock \showarticletitle{End-to-end object detection with transformers}. In
  \bibinfo{booktitle}{\emph{Proceedings of the European Conference Computer
  Vision}}. Springer, \bibinfo{pages}{213--229}.
\newblock


\bibitem[Chai et~al\mbox{.}(2022)]%
        {chai2022pdd}
\bibfield{author}{\bibinfo{person}{Xiaoxuan Chai}, \bibinfo{person}{Junchi
  Zhou}, \bibinfo{person}{Hang Zhou}, {and} \bibinfo{person}{Juihsin Lai}.}
  \bibinfo{year}{2022}\natexlab{}.
\newblock \showarticletitle{PDD-GAN: Prior-based GAN Network with Decoupling
  Ability for Single Image Dehazing}. In \bibinfo{booktitle}{\emph{Proceedings
  of the 30th ACM International Conference on Multimedia}}.
  \bibinfo{pages}{5952--5960}.
\newblock


\bibitem[Chen et~al\mbox{.}(2022a)]%
        {chen2022simple}
\bibfield{author}{\bibinfo{person}{Liangyu Chen}, \bibinfo{person}{Xiaojie
  Chu}, \bibinfo{person}{Xiangyu Zhang}, {and} \bibinfo{person}{Jian Sun}.}
  \bibinfo{year}{2022}\natexlab{a}.
\newblock \showarticletitle{Simple baselines for image restoration}. In
  \bibinfo{booktitle}{\emph{Proceedings of the European Conference Computer
  Vision}}. \bibinfo{pages}{17--33}.
\newblock


\bibitem[Chen et~al\mbox{.}(2021a)]%
        {chen2021hinet}
\bibfield{author}{\bibinfo{person}{Liangyu Chen}, \bibinfo{person}{Xin Lu},
  \bibinfo{person}{Jie Zhang}, \bibinfo{person}{Xiaojie Chu}, {and}
  \bibinfo{person}{Chengpeng Chen}.} \bibinfo{year}{2021}\natexlab{a}.
\newblock \showarticletitle{HINet: Half instance normalization network for
  image restoration}. In \bibinfo{booktitle}{\emph{Proceedings of the IEEE/CVF
  Conference on Computer Vision and Pattern Recognition Workshops}}.
  \bibinfo{pages}{182--192}.
\newblock


\bibitem[Chen et~al\mbox{.}(2022b)]%
        {chen2022snowformer}
\bibfield{author}{\bibinfo{person}{Sixiang Chen}, \bibinfo{person}{Tian Ye},
  \bibinfo{person}{Yun Liu}, \bibinfo{person}{Erkang Chen},
  \bibinfo{person}{Jun Shi}, {and} \bibinfo{person}{Jingchun Zhou}.}
  \bibinfo{year}{2022}\natexlab{b}.
\newblock \showarticletitle{SnowFormer: scale-aware transformer via context
  interaction for single image desnowing}.
\newblock \bibinfo{journal}{\emph{arXiv preprint arXiv:2208.09703}}
  (\bibinfo{year}{2022}).
\newblock


\bibitem[Chen et~al\mbox{.}(2023a)]%
        {chen2023msp}
\bibfield{author}{\bibinfo{person}{Sixiang Chen}, \bibinfo{person}{Tian Ye},
  \bibinfo{person}{Yun Liu}, \bibinfo{person}{Taodong Liao},
  \bibinfo{person}{Jingxia Jiang}, \bibinfo{person}{Erkang Chen}, {and}
  \bibinfo{person}{Peng Chen}.} \bibinfo{year}{2023}\natexlab{a}.
\newblock \showarticletitle{MSP-former: Multi-scale projection transformer for
  single image desnowing}. In \bibinfo{booktitle}{\emph{Proceedings of the IEEE
  International Conference on Acoustics, Speech and Signal Processing
  (ICASSP)}}. \bibinfo{pages}{1--5}.
\newblock
\urldef\tempurl%
\url{https://doi.org/10.1109/ICASSP49357.2023.10095605}
\showDOI{\tempurl}


\bibitem[Chen et~al\mbox{.}(2023b)]%
        {chen2023dehrformer}
\bibfield{author}{\bibinfo{person}{Sixiang Chen}, \bibinfo{person}{Tian Ye},
  \bibinfo{person}{Jun Shi}, \bibinfo{person}{Yun Liu},
  \bibinfo{person}{JingXia Jiang}, \bibinfo{person}{Erkang Chen}, {and}
  \bibinfo{person}{Peng Chen}.} \bibinfo{year}{2023}\natexlab{b}.
\newblock \showarticletitle{DEHRFormer: Real-time transformer for depth
  estimation and haze removal from varicolored haze scenes}. In
  \bibinfo{booktitle}{\emph{Proceedings of the IEEE International Conference on
  Acoustics, Speech and Signal Processing (ICASSP)}}. \bibinfo{pages}{1--5}.
\newblock
\urldef\tempurl%
\url{https://doi.org/10.1109/ICASSP49357.2023.10096828}
\showDOI{\tempurl}


\bibitem[Chen et~al\mbox{.}(2021b)]%
        {chen2021psd}
\bibfield{author}{\bibinfo{person}{Zeyuan Chen}, \bibinfo{person}{Yangchao
  Wang}, \bibinfo{person}{Yang Yang}, {and} \bibinfo{person}{Dong Liu}.}
  \bibinfo{year}{2021}\natexlab{b}.
\newblock \showarticletitle{PSD: Principled synthetic-to-real dehazing guided
  by physical priors}. In \bibinfo{booktitle}{\emph{Proceedings of the IEEE/CVF
  Conference on Computer Vision and Pattern Recognition}}.
  \bibinfo{pages}{7180--7189}.
\newblock


\bibitem[Dosovitskiy et~al\mbox{.}(2020)]%
        {dosovitskiy2020image}
\bibfield{author}{\bibinfo{person}{Alexey Dosovitskiy}, \bibinfo{person}{Lucas
  Beyer}, \bibinfo{person}{Alexander Kolesnikov}, \bibinfo{person}{Dirk
  Weissenborn}, \bibinfo{person}{Xiaohua Zhai}, \bibinfo{person}{Thomas
  Unterthiner}, \bibinfo{person}{Mostafa Dehghani}, \bibinfo{person}{Matthias
  Minderer}, \bibinfo{person}{Georg Heigold}, \bibinfo{person}{Sylvain Gelly},
  {et~al\mbox{.}}} \bibinfo{year}{2020}\natexlab{}.
\newblock \showarticletitle{An image is worth 16x16 words: Transformers for
  image recognition at scale}.
\newblock \bibinfo{journal}{\emph{arXiv preprint arXiv:2010.11929}}
  (\bibinfo{year}{2020}).
\newblock


\bibitem[{Epic Games, Inc.}(2021)]%
        {UE}
\bibfield{author}{\bibinfo{person}{{Epic Games, Inc.}}}
  \bibinfo{year}{2021}\natexlab{}.
\newblock \bibinfo{booktitle}{\emph{Unreal Engine}}.
\newblock
\urldef\tempurl%
\url{https://www.unrealengine.com/}
\showURL{%
\tempurl}


\bibitem[Fattal(2014)]%
        {fattal2014dehazing}
\bibfield{author}{\bibinfo{person}{Raanan Fattal}.}
  \bibinfo{year}{2014}\natexlab{}.
\newblock \showarticletitle{Dehazing using color-lines}.
\newblock \bibinfo{journal}{\emph{ACM Transactions on Graphics}}
  \bibinfo{volume}{34}, \bibinfo{number}{1} (\bibinfo{year}{2014}),
  \bibinfo{pages}{1--14}.
\newblock


\bibitem[Guo et~al\mbox{.}(2020)]%
        {guo2020zero}
\bibfield{author}{\bibinfo{person}{Chunle Guo}, \bibinfo{person}{Chongyi Li},
  \bibinfo{person}{Jichang Guo}, \bibinfo{person}{Chen~Change Loy},
  \bibinfo{person}{Junhui Hou}, \bibinfo{person}{Sam Kwong}, {and}
  \bibinfo{person}{Runmin Cong}.} \bibinfo{year}{2020}\natexlab{}.
\newblock \showarticletitle{Zero-reference deep curve estimation for low-light
  image enhancement}. In \bibinfo{booktitle}{\emph{Proceedings of the IEEE/CVF
  Conference on Computer Vision and Pattern Recognition}}.
  \bibinfo{pages}{1780--1789}.
\newblock


\bibitem[He et~al\mbox{.}(2011)]%
        {he2010single}
\bibfield{author}{\bibinfo{person}{Kaiming He}, \bibinfo{person}{Jian Sun},
  {and} \bibinfo{person}{Xiaoou Tang}.} \bibinfo{year}{2011}\natexlab{}.
\newblock \showarticletitle{Single image haze removal using dark channel
  prior}.
\newblock \bibinfo{journal}{\emph{IEEE Transactions on Pattern Analysis and
  Machine Intelligence}} \bibinfo{volume}{33}, \bibinfo{number}{12}
  (\bibinfo{year}{2011}), \bibinfo{pages}{2341--2353}.
\newblock


\bibitem[He et~al\mbox{.}(2016)]%
        {he2016deep}
\bibfield{author}{\bibinfo{person}{Kaiming He}, \bibinfo{person}{Xiangyu
  Zhang}, \bibinfo{person}{Shaoqing Ren}, {and} \bibinfo{person}{Jian Sun}.}
  \bibinfo{year}{2016}\natexlab{}.
\newblock \showarticletitle{Deep residual learning for image recognition}. In
  \bibinfo{booktitle}{\emph{Proceedings of the IEEE Conference on Computer
  Vision and Pattern Recognition}}. \bibinfo{pages}{770--778}.
\newblock


\bibitem[Heusel et~al\mbox{.}(2017)]%
        {heusel2017gans}
\bibfield{author}{\bibinfo{person}{Martin Heusel}, \bibinfo{person}{Hubert
  Ramsauer}, \bibinfo{person}{Thomas Unterthiner}, \bibinfo{person}{Bernhard
  Nessler}, {and} \bibinfo{person}{Sepp Hochreiter}.}
  \bibinfo{year}{2017}\natexlab{}.
\newblock \showarticletitle{Gans trained by a two time-scale update rule
  converge to a local nash equilibrium}. In \bibinfo{booktitle}{\emph{Advances
  in Neural Information Processing Systems}}. \bibinfo{pages}{1--12}.
\newblock


\bibitem[Huang et~al\mbox{.}(2022a)]%
        {Huang_2022_CVPR}
\bibfield{author}{\bibinfo{person}{Jie Huang}, \bibinfo{person}{Yajing Liu},
  \bibinfo{person}{Xueyang Fu}, \bibinfo{person}{Man Zhou},
  \bibinfo{person}{Yang Wang}, \bibinfo{person}{Feng Zhao}, {and}
  \bibinfo{person}{Zhiwei Xiong}.} \bibinfo{year}{2022}\natexlab{a}.
\newblock \showarticletitle{Exposure normalization and compensation for
  multiple-exposure correction}. In \bibinfo{booktitle}{\emph{Proceedings of
  the IEEE/CVF Conference on Computer Vision and Pattern Recognition}}.
  \bibinfo{pages}{6043--6052}.
\newblock


\bibitem[Huang et~al\mbox{.}(2022b)]%
        {huang2022deep}
\bibfield{author}{\bibinfo{person}{Jie Huang}, \bibinfo{person}{Yajing Liu},
  \bibinfo{person}{Feng Zhao}, \bibinfo{person}{Keyu Yan},
  \bibinfo{person}{Jinghao Zhang}, \bibinfo{person}{Yukun Huang},
  \bibinfo{person}{Man Zhou}, {and} \bibinfo{person}{Zhiwei Xiong}.}
  \bibinfo{year}{2022}\natexlab{b}.
\newblock \showarticletitle{Deep fourier-based exposure correction network with
  spatial-frequency interaction}. In \bibinfo{booktitle}{\emph{Proceedings of
  the European Conference Computer Vision}}. \bibinfo{pages}{163--180}.
\newblock


\bibitem[Huang et~al\mbox{.}(2019)]%
        {HPEU}
\bibfield{author}{\bibinfo{person}{Jie Huang}, \bibinfo{person}{Zhiwei Xiong},
  \bibinfo{person}{Xueyang Fu}, \bibinfo{person}{Dong Liu}, {and}
  \bibinfo{person}{Zheng-Jun Zha}.} \bibinfo{year}{2019}\natexlab{}.
\newblock \showarticletitle{Hybrid image enhancement with progressive laplacian
  enhancing unit}. In \bibinfo{booktitle}{\emph{Proceedings of the 27th ACM
  International Conference on Multimedia}}. \bibinfo{pages}{1614--1622}.
\newblock


\bibitem[Huang et~al\mbox{.}(2023)]%
        {Huang_2023_CVPR}
\bibfield{author}{\bibinfo{person}{Jie Huang}, \bibinfo{person}{Feng Zhao},
  \bibinfo{person}{Man Zhou}, \bibinfo{person}{Jie Xiao},
  \bibinfo{person}{Naishan Zheng}, \bibinfo{person}{Kaiwen Zheng}, {and}
  \bibinfo{person}{Zhiwei Xiong}.} \bibinfo{year}{2023}\natexlab{}.
\newblock \showarticletitle{Learning sample relationship for exposure
  correction}. In \bibinfo{booktitle}{\emph{Proceedings of the IEEE/CVF
  Conference on Computer Vision and Pattern Recognition}}.
  \bibinfo{pages}{9904--9913}.
\newblock


\bibitem[Huang et~al\mbox{.}(2022c)]%
        {ECLNet}
\bibfield{author}{\bibinfo{person}{Jie Huang}, \bibinfo{person}{Man Zhou},
  \bibinfo{person}{Yajing Liu}, \bibinfo{person}{Mingde Yao},
  \bibinfo{person}{Feng Zhao}, {and} \bibinfo{person}{Zhiwei Xiong}.}
  \bibinfo{year}{2022}\natexlab{c}.
\newblock \showarticletitle{Exposure-consistency representation learning for
  exposure correction}. In \bibinfo{booktitle}{\emph{Proceedings of the 30th
  ACM International Conference on Multimedia}}. \bibinfo{pages}{6309--6317}.
\newblock


\bibitem[Huang et~al\mbox{.}(2018)]%
        {Huang_2018_ECCV_Workshops}
\bibfield{author}{\bibinfo{person}{Jie Huang}, \bibinfo{person}{Pengfei Zhu},
  \bibinfo{person}{Mingrui Geng}, \bibinfo{person}{Jiewen Ran},
  \bibinfo{person}{Xingguang Zhou}, \bibinfo{person}{Chen Xing},
  \bibinfo{person}{Pengfei Wan}, {and} \bibinfo{person}{Xiangyang Ji}.}
  \bibinfo{year}{2018}\natexlab{}.
\newblock \showarticletitle{Range scaling global u-net for perceptual image
  enhancement on mobile devices}. In \bibinfo{booktitle}{\emph{Proceedings of
  the European Conference on Computer Vision Workshops}}.
  \bibinfo{pages}{230--242}.
\newblock


\bibitem[Jiang et~al\mbox{.}(2023)]%
        {jiang2023rsfdm}
\bibfield{author}{\bibinfo{person}{Jingxia Jiang}, \bibinfo{person}{Jinbin
  Bai}, \bibinfo{person}{Yun Liu}, \bibinfo{person}{Junjie Yin},
  \bibinfo{person}{Sixiang Chen}, \bibinfo{person}{Tian Ye}, {and}
  \bibinfo{person}{Erkang Chen}.} \bibinfo{year}{2023}\natexlab{}.
\newblock \showarticletitle{RSFDM-Net: Real-time spatial and frequency domains
  modulation network for underwater image enhancement}.
\newblock \bibinfo{journal}{\emph{arXiv preprint arXiv:2302.12186}}
  (\bibinfo{year}{2023}).
\newblock


\bibitem[Jin et~al\mbox{.}(2023)]%
        {jin2023enhancing}
\bibfield{author}{\bibinfo{person}{Yeying Jin}, \bibinfo{person}{Beibei Lin},
  \bibinfo{person}{Wending Yan}, \bibinfo{person}{Wei Ye},
  \bibinfo{person}{Yuan Yuan}, {and} \bibinfo{person}{Robby~T. Tan}.}
  \bibinfo{year}{2023}\natexlab{}.
\newblock \showarticletitle{Enhancing visibility in nighttime haze images using
  guided APSF and gradient adaptive convolution}.
\newblock \bibinfo{journal}{\emph{arXiv preprint arXiv:2308.01738}}
  (\bibinfo{year}{2023}).
\newblock


\bibitem[Ju et~al\mbox{.}(2021)]%
        {ju2021idrlp}
\bibfield{author}{\bibinfo{person}{Mingye. Ju}, \bibinfo{person}{Can Ding},
  \bibinfo{person}{Charles~A. Guo}, \bibinfo{person}{Wenqi Ren}, {and}
  \bibinfo{person}{Dacheng Tao}.} \bibinfo{year}{2021}\natexlab{}.
\newblock \showarticletitle{IDRLP: Image dehazing using region line prior}.
\newblock \bibinfo{journal}{\emph{IEEE Transactions on Image Processing}}
  \bibinfo{volume}{30} (\bibinfo{year}{2021}), \bibinfo{pages}{9043--9057}.
\newblock


\bibitem[Ju et~al\mbox{.}(2019)]%
        {ju2019idgcp}
\bibfield{author}{\bibinfo{person}{Mingye Ju}, \bibinfo{person}{Can Ding},
  \bibinfo{person}{Y.~Jay Guo}, {and} \bibinfo{person}{Dengyin Zhang}.}
  \bibinfo{year}{2019}\natexlab{}.
\newblock \showarticletitle{IDGCP: Image dehazing based on gamma correction
  prior}.
\newblock \bibinfo{journal}{\emph{IEEE Transactions on Image Processing}}
  \bibinfo{volume}{29} (\bibinfo{year}{2019}), \bibinfo{pages}{3104--3118}.
\newblock


\bibitem[Ke et~al\mbox{.}(2021)]%
        {ke2021musiq}
\bibfield{author}{\bibinfo{person}{Junjie Ke}, \bibinfo{person}{Qifei Wang},
  \bibinfo{person}{Yilin Wang}, \bibinfo{person}{Peyman Milanfar}, {and}
  \bibinfo{person}{Feng Yang}.} \bibinfo{year}{2021}\natexlab{}.
\newblock \showarticletitle{MUSIQ: Multi-scale image quality transformer}. In
  \bibinfo{booktitle}{\emph{Proceedings of the IEEE/CVF International
  Conference on Computer Vision}}. \bibinfo{pages}{5148--5157}.
\newblock


\bibitem[Koo and Kim(2020)]%
        {koo2019nighttime}
\bibfield{author}{\bibinfo{person}{Beomhyuk Koo} {and}
  \bibinfo{person}{Gyeonghwan Kim}.} \bibinfo{year}{2020}\natexlab{}.
\newblock \showarticletitle{Nighttime haze removal with glow decomposition
  using GAN}. In \bibinfo{booktitle}{\emph{Proceedings of the Asian Conference
  Pattern Recognition}}. \bibinfo{pages}{807--820}.
\newblock


\bibitem[Kuanar et~al\mbox{.}(2022)]%
        {kuanar2022multi}
\bibfield{author}{\bibinfo{person}{Shiba Kuanar}, \bibinfo{person}{Dwarikanath
  Mahapatra}, \bibinfo{person}{Monalisa Bilas}, {and} \bibinfo{person}{K.~R.
  Rao}.} \bibinfo{year}{2022}\natexlab{}.
\newblock \showarticletitle{Multi-path dilated convolution network for haze and
  glow removal in nighttime images}.
\newblock \bibinfo{journal}{\emph{The Visual Computer}}  \bibinfo{volume}{38}
  (\bibinfo{year}{2022}), \bibinfo{pages}{1121--1134}.
\newblock


\bibitem[Li et~al\mbox{.}(2017)]%
        {li2017aod}
\bibfield{author}{\bibinfo{person}{Boyi Li}, \bibinfo{person}{Xiulian Peng},
  \bibinfo{person}{Zhangyang Wang}, \bibinfo{person}{Jizheng Xu}, {and}
  \bibinfo{person}{Dan Feng}.} \bibinfo{year}{2017}\natexlab{}.
\newblock \showarticletitle{AOD-Net: All-in-one dehazing network}. In
  \bibinfo{booktitle}{\emph{Proceedings of the IEEE International Conference on
  Computer Vision}}. \bibinfo{pages}{4770--4778}.
\newblock


\bibitem[Li et~al\mbox{.}(2015)]%
        {li2015nighttime}
\bibfield{author}{\bibinfo{person}{Yu Li}, \bibinfo{person}{Robby~T. Tan},
  {and} \bibinfo{person}{Michael~S. Brown}.} \bibinfo{year}{2015}\natexlab{}.
\newblock \showarticletitle{Nighttime haze removal with glow and multiple light
  colors}. In \bibinfo{booktitle}{\emph{Proceedings of the IEEE International
  Conference on Computer Vision}}. \bibinfo{pages}{226--234}.
\newblock


\bibitem[Liao et~al\mbox{.}(2018)]%
        {liao2018hdp}
\bibfield{author}{\bibinfo{person}{Yinghong Liao}, \bibinfo{person}{Zhuo Su},
  \bibinfo{person}{Xiangguo Liang}, {and} \bibinfo{person}{Bin Qiu}.}
  \bibinfo{year}{2018}\natexlab{}.
\newblock \showarticletitle{Hdp-net: Haze density prediction network for
  nighttime dehazing}. In \bibinfo{booktitle}{\emph{Proceedings of the
  Pacific-Rim Conference on Multimedia}}. \bibinfo{pages}{469--480}.
\newblock


\bibitem[Liu et~al\mbox{.}(2021)]%
        {liu2021single}
\bibfield{author}{\bibinfo{person}{Yun Liu}, \bibinfo{person}{Anzhi Wang},
  \bibinfo{person}{Hao Zhou}, {and} \bibinfo{person}{Pengfei Jia}.}
  \bibinfo{year}{2021}\natexlab{}.
\newblock \showarticletitle{Single nighttime image dehazing based on image
  decomposition}.
\newblock \bibinfo{journal}{\emph{Signal Processing}}  \bibinfo{volume}{183}
  (\bibinfo{year}{2021}), \bibinfo{pages}{107986}.
\newblock


\bibitem[Liu et~al\mbox{.}(2023)]%
        {liu2023multi}
\bibfield{author}{\bibinfo{person}{Yun Liu}, \bibinfo{person}{Zhongsheng Yan},
  \bibinfo{person}{Jinge Tan}, {and} \bibinfo{person}{Yuche Li}.}
  \bibinfo{year}{2023}\natexlab{}.
\newblock \showarticletitle{Multi-purpose oriented single nighttime image haze
  removal based on unified variational retinex model}.
\newblock \bibinfo{journal}{\emph{IEEE Transactions on Circuits and Systems for
  Video Technology}} \bibinfo{volume}{33}, \bibinfo{number}{4}
  (\bibinfo{year}{2023}), \bibinfo{pages}{1643--1657}.
\newblock


\bibitem[Liu et~al\mbox{.}(2022a)]%
        {liu2022nighttime}
\bibfield{author}{\bibinfo{person}{Yun Liu}, \bibinfo{person}{Zhongsheng Yan},
  \bibinfo{person}{Aimin Wu}, \bibinfo{person}{Tian Ye}, {and}
  \bibinfo{person}{Yuche Li}.} \bibinfo{year}{2022}\natexlab{a}.
\newblock \showarticletitle{Nighttime image dehazing based on variational
  decomposition model}. In \bibinfo{booktitle}{\emph{Proceedings of the
  IEEE/CVF Conference on Computer Vision and Pattern Recognition Workshops}}.
  \bibinfo{pages}{640--649}.
\newblock


\bibitem[Liu et~al\mbox{.}(2022b)]%
        {liu2022single}
\bibfield{author}{\bibinfo{person}{Yun Liu}, \bibinfo{person}{Zhongsheng Yan},
  \bibinfo{person}{Tian Ye}, \bibinfo{person}{Aimin Wu}, {and}
  \bibinfo{person}{Yuche Li}.} \bibinfo{year}{2022}\natexlab{b}.
\newblock \showarticletitle{Single nighttime image dehazing based on unified
  variational decomposition model and multi-scale contrast enhancement}.
\newblock \bibinfo{journal}{\emph{Engineering Applications of Artificial
  Intelligence}}  \bibinfo{volume}{116} (\bibinfo{year}{2022}),
  \bibinfo{pages}{105373}.
\newblock


\bibitem[McCartney(1976)]%
        {mccartney1976optics}
\bibfield{author}{\bibinfo{person}{E.~J. McCartney}.}
  \bibinfo{year}{1976}\natexlab{}.
\newblock \showarticletitle{Optics of the atmosphere: scattering by molecules
  and particles}.
\newblock \bibinfo{journal}{\emph{New York}} (\bibinfo{year}{1976}).
\newblock


\bibitem[Meng et~al\mbox{.}(2013)]%
        {meng2013efficient}
\bibfield{author}{\bibinfo{person}{Gaofeng Meng}, \bibinfo{person}{Ying Wang},
  \bibinfo{person}{Jiangyong Duan}, \bibinfo{person}{Shiming Xiang}, {and}
  \bibinfo{person}{Chunhong Pan}.} \bibinfo{year}{2013}\natexlab{}.
\newblock \showarticletitle{Efficient image dehazing with boundary constraint
  and contextual regularization}. In \bibinfo{booktitle}{\emph{Proceedings of
  the IEEE International Conference on Computer Vision}}.
  \bibinfo{pages}{617--624}.
\newblock


\bibitem[Mittal et~al\mbox{.}(2013)]%
        {mittal2012making}
\bibfield{author}{\bibinfo{person}{Anish Mittal}, \bibinfo{person}{Rajiv
  Soundararajan}, {and} \bibinfo{person}{Alan~C. Bovik}.}
  \bibinfo{year}{2013}\natexlab{}.
\newblock \showarticletitle{Making a ``completely blind'' image quality
  analyzer}.
\newblock \bibinfo{journal}{\emph{IEEE Signal Processing Letters}}
  \bibinfo{volume}{20}, \bibinfo{number}{3} (\bibinfo{year}{2013}),
  \bibinfo{pages}{209--212}.
\newblock


\bibitem[Murray et~al\mbox{.}(2012)]%
        {murray2012ava}
\bibfield{author}{\bibinfo{person}{Naila Murray}, \bibinfo{person}{Luca
  Marchesotti}, {and} \bibinfo{person}{Florent Perronnin}.}
  \bibinfo{year}{2012}\natexlab{}.
\newblock \showarticletitle{AVA: A large-scale database for aesthetic visual
  analysis}. In \bibinfo{booktitle}{\emph{Proceedings of IEEE Conference on
  Computer Vision and Pattern Recognition}}. \bibinfo{pages}{2408--2415}.
\newblock


\bibitem[Paszke et~al\mbox{.}(2019)]%
        {paszke2019pytorch}
\bibfield{author}{\bibinfo{person}{Adam Paszke}, \bibinfo{person}{Sam Gross},
  \bibinfo{person}{Francisco Massa}, \bibinfo{person}{Adam Lerer},
  \bibinfo{person}{James Bradbury}, \bibinfo{person}{Gregory Chanan},
  \bibinfo{person}{Trevor Killeen}, \bibinfo{person}{Zeming Lin},
  \bibinfo{person}{Natalia Gimelshein}, \bibinfo{person}{Luca Antiga},
  {et~al\mbox{.}}} \bibinfo{year}{2019}\natexlab{}.
\newblock \showarticletitle{Pytorch: An imperative style, high-performance deep
  learning library}. In \bibinfo{booktitle}{\emph{Advances in Neural
  Information Processing Systems}}, Vol.~\bibinfo{volume}{32}.
\newblock


\bibitem[Qin et~al\mbox{.}(2020)]%
        {qin2020ffa}
\bibfield{author}{\bibinfo{person}{Xu Qin}, \bibinfo{person}{Zhilin Wang},
  \bibinfo{person}{Yuanchao Bai}, \bibinfo{person}{Xiaodong Xie}, {and}
  \bibinfo{person}{Huizhu Jia}.} \bibinfo{year}{2020}\natexlab{}.
\newblock \showarticletitle{FFA-Net: Feature fusion attention network for
  single image dehazing}. In \bibinfo{booktitle}{\emph{Proceedings of the AAAI
  Conference on Artificial Intelligence}}, Vol.~\bibinfo{volume}{34}.
  \bibinfo{pages}{11908--11915}.
\newblock


\bibitem[Ren et~al\mbox{.}(2016)]%
        {ren2016single}
\bibfield{author}{\bibinfo{person}{Wenqi Ren}, \bibinfo{person}{Si Liu},
  \bibinfo{person}{Hua Zhang}, \bibinfo{person}{Jinshan Pan},
  \bibinfo{person}{Xiaochun Cao}, {and} \bibinfo{person}{Ming-Hsuan Yang}.}
  \bibinfo{year}{2016}\natexlab{}.
\newblock \showarticletitle{Single image dehazing via multi-scale convolutional
  neural networks}. In \bibinfo{booktitle}{\emph{Proceedings of the European
  Conference Computer Vision}}. \bibinfo{pages}{154--169}.
\newblock


\bibitem[Simonyan and Zisserman(2014)]%
        {simonyan2014very}
\bibfield{author}{\bibinfo{person}{Karen Simonyan} {and}
  \bibinfo{person}{Andrew Zisserman}.} \bibinfo{year}{2014}\natexlab{}.
\newblock \showarticletitle{Very deep convolutional networks for large-scale
  image recognition}.
\newblock \bibinfo{journal}{\emph{arXiv preprint arXiv:1409.1556}}
  (\bibinfo{year}{2014}).
\newblock


\bibitem[Valanarasu et~al\mbox{.}(2022)]%
        {valanarasu2022transweather}
\bibfield{author}{\bibinfo{person}{Jeya Maria~Jose Valanarasu},
  \bibinfo{person}{Rajeev Yasarla}, {and} \bibinfo{person}{Vishal~M. Patel}.}
  \bibinfo{year}{2022}\natexlab{}.
\newblock \showarticletitle{Transweather: Transformer-based restoration of
  images degraded by adverse weather conditions}. In
  \bibinfo{booktitle}{\emph{Proceedings of the IEEE/CVF Conference on Computer
  Vision and Pattern Recognition}}. \bibinfo{pages}{2353--2363}.
\newblock


\bibitem[Van~der Maaten and Hinton(2008)]%
        {van2008visualizing}
\bibfield{author}{\bibinfo{person}{Laurens Van~der Maaten} {and}
  \bibinfo{person}{Geoffrey Hinton}.} \bibinfo{year}{2008}\natexlab{}.
\newblock \showarticletitle{Visualizing data using t-SNE.}
\newblock \bibinfo{journal}{\emph{Journal of Machine Learning Research}}
  \bibinfo{volume}{9} (\bibinfo{year}{2008}), \bibinfo{pages}{2579--2605}.
\newblock


\bibitem[Wang et~al\mbox{.}(2022a)]%
        {wang2022variational}
\bibfield{author}{\bibinfo{person}{Wenhui Wang}, \bibinfo{person}{Anna Wang},
  {and} \bibinfo{person}{Chen Liu}.} \bibinfo{year}{2022}\natexlab{a}.
\newblock \showarticletitle{Variational single nighttime image haze removal
  with a gray haze-line prior}.
\newblock \bibinfo{journal}{\emph{IEEE Transactions on Image Processing}}
  \bibinfo{volume}{31} (\bibinfo{year}{2022}), \bibinfo{pages}{1349--1363}.
\newblock


\bibitem[Wang et~al\mbox{.}(2022b)]%
        {wang2022rapid}
\bibfield{author}{\bibinfo{person}{Wenhui Wang}, \bibinfo{person}{Anna Wang},
  \bibinfo{person}{Xingyu Wang}, \bibinfo{person}{Haijing Sun}, {and}
  \bibinfo{person}{Qing Ai}.} \bibinfo{year}{2022}\natexlab{b}.
\newblock \showarticletitle{Rapid nighttime haze removal with color-gray layer
  decomposition}.
\newblock \bibinfo{journal}{\emph{Signal Processing}}  \bibinfo{volume}{200}
  (\bibinfo{year}{2022}), \bibinfo{pages}{108658}.
\newblock


\bibitem[Wang et~al\mbox{.}(2013)]%
        {wang2013automatic}
\bibfield{author}{\bibinfo{person}{Yinting Wang}, \bibinfo{person}{Shaojie
  Zhuo}, \bibinfo{person}{Dapeng Tao}, \bibinfo{person}{Jiajun Bu}, {and}
  \bibinfo{person}{Na Li}.} \bibinfo{year}{2013}\natexlab{}.
\newblock \showarticletitle{Automatic local exposure correction using bright
  channel prior for under-exposed images}.
\newblock \bibinfo{journal}{\emph{Signal Processing}} \bibinfo{volume}{93},
  \bibinfo{number}{11} (\bibinfo{year}{2013}), \bibinfo{pages}{3227--3238}.
\newblock


\bibitem[Wang et~al\mbox{.}(2004)]%
        {wang2004image}
\bibfield{author}{\bibinfo{person}{Zhou Wang}, \bibinfo{person}{Alan~C. Bovik},
  \bibinfo{person}{Hamid~R. Sheikh}, {and} \bibinfo{person}{Eero~P.
  Simoncelli}.} \bibinfo{year}{2004}\natexlab{}.
\newblock \showarticletitle{Image quality assessment: from error visibility to
  structural similarity}.
\newblock \bibinfo{journal}{\emph{IEEE Transactions on Image Processing}}
  \bibinfo{volume}{13}, \bibinfo{number}{4} (\bibinfo{year}{2004}),
  \bibinfo{pages}{600--612}.
\newblock


\bibitem[Wu et~al\mbox{.}(2021)]%
        {wu2021contrastive}
\bibfield{author}{\bibinfo{person}{Haiyan Wu}, \bibinfo{person}{Yanyun Qu},
  \bibinfo{person}{Shaohui Lin}, \bibinfo{person}{Jian Zhou},
  \bibinfo{person}{Ruizhi Qiao}, \bibinfo{person}{Zhizhong Zhang},
  \bibinfo{person}{Yuan Xie}, {and} \bibinfo{person}{Lizhuang Ma}.}
  \bibinfo{year}{2021}\natexlab{}.
\newblock \showarticletitle{Contrastive learning for compact single image
  dehazing}. In \bibinfo{booktitle}{\emph{Proceedings of the IEEE/CVF
  Conference on Computer Vision and Pattern Recognition}}.
  \bibinfo{pages}{10551--10560}.
\newblock


\bibitem[Wu et~al\mbox{.}(2023)]%
        {wu2023ridcp}
\bibfield{author}{\bibinfo{person}{Rui-Qi Wu}, \bibinfo{person}{Zheng-Peng
  Duan}, \bibinfo{person}{Chun-Le Guo}, \bibinfo{person}{Zhi Chai}, {and}
  \bibinfo{person}{Chongyi Li}.} \bibinfo{year}{2023}\natexlab{}.
\newblock \showarticletitle{RIDCP: Revitalizing real image dehazing via
  high-quality codebook priors}.
\newblock \bibinfo{journal}{\emph{Proceedings of the IEEE/CVF Conference on
  Computer Vision and Pattern Recognition}} (\bibinfo{year}{2023}),
  \bibinfo{pages}{22282--22291}.
\newblock


\bibitem[Yan et~al\mbox{.}(2020)]%
        {yan2020nighttime}
\bibfield{author}{\bibinfo{person}{Wending Yan}, \bibinfo{person}{Robby~T Tan},
  {and} \bibinfo{person}{Dengxin Dai}.} \bibinfo{year}{2020}\natexlab{}.
\newblock \showarticletitle{Nighttime defogging using high-low frequency
  decomposition and grayscale-color networks}. In
  \bibinfo{booktitle}{\emph{Proceedings of the European Conference Computer
  Vision}}. Springer, \bibinfo{pages}{473--488}.
\newblock


\bibitem[Yang et~al\mbox{.}(2023)]%
        {Yang_2023_CVPR}
\bibfield{author}{\bibinfo{person}{Zizheng Yang}, \bibinfo{person}{Jie Huang},
  \bibinfo{person}{Jiahao Chang}, \bibinfo{person}{Man Zhou},
  \bibinfo{person}{Hu Yu}, \bibinfo{person}{Jinghao Zhang}, {and}
  \bibinfo{person}{Feng Zhao}.} \bibinfo{year}{2023}\natexlab{}.
\newblock \showarticletitle{Visual recognition-driven image restoration for
  multiple degradation with intrinsic semantics recovery}. In
  \bibinfo{booktitle}{\emph{Proceedings of the IEEE/CVF Conference on Computer
  Vision and Pattern Recognition}}. \bibinfo{pages}{14059--14070}.
\newblock


\bibitem[Ye et~al\mbox{.}(2022a)]%
        {ye2022towards}
\bibfield{author}{\bibinfo{person}{Tian Ye}, \bibinfo{person}{Sixiang Chen},
  \bibinfo{person}{Yun Liu}, \bibinfo{person}{Yi Ye}, \bibinfo{person}{Jinbin
  Bai}, {and} \bibinfo{person}{Erkang Chen}.} \bibinfo{year}{2022}\natexlab{a}.
\newblock \showarticletitle{Towards real-time high-definition image snow
  removal: Efficient pyramid network with asymmetrical encoder-decoder
  architecture}. In \bibinfo{booktitle}{\emph{Proceedings of the Asian
  Conference on Computer Vision}}. \bibinfo{pages}{366--381}.
\newblock


\bibitem[Ye et~al\mbox{.}(2022b)]%
        {ye2022underwater}
\bibfield{author}{\bibinfo{person}{Tian Ye}, \bibinfo{person}{Sixiang Chen},
  \bibinfo{person}{Yun Liu}, \bibinfo{person}{Yi Ye}, \bibinfo{person}{Erkang
  Chen}, {and} \bibinfo{person}{Yuche Li}.} \bibinfo{year}{2022}\natexlab{b}.
\newblock \showarticletitle{Underwater light field retention: Neural rendering
  for underwater imaging}. In \bibinfo{booktitle}{\emph{Proceedings of the
  IEEE/CVF Conference on Computer Vision and Pattern Recognition Workshops}}.
  \bibinfo{pages}{488--497}.
\newblock


\bibitem[Ye et~al\mbox{.}(2022c)]%
        {ye2022perceiving}
\bibfield{author}{\bibinfo{person}{Tian Ye}, \bibinfo{person}{Yunchen Zhang},
  \bibinfo{person}{Mingchao Jiang}, \bibinfo{person}{Liang Chen},
  \bibinfo{person}{Yun Liu}, \bibinfo{person}{Sixiang Chen}, {and}
  \bibinfo{person}{Erkang Chen}.} \bibinfo{year}{2022}\natexlab{c}.
\newblock \showarticletitle{Perceiving and Modeling Density for Image
  Dehazing}. In \bibinfo{booktitle}{\emph{Proceedings of the European
  Conference Computer Vision}}. Springer, \bibinfo{pages}{130--145}.
\newblock


\bibitem[Yu et~al\mbox{.}(2022)]%
        {yu2022source}
\bibfield{author}{\bibinfo{person}{Hu Yu}, \bibinfo{person}{Jie Huang},
  \bibinfo{person}{Yajing Liu}, \bibinfo{person}{Qi Zhu}, \bibinfo{person}{Man
  Zhou}, {and} \bibinfo{person}{Feng Zhao}.} \bibinfo{year}{2022}\natexlab{}.
\newblock \showarticletitle{Source-free domain adaptation for real-world image
  dehazing}. In \bibinfo{booktitle}{\emph{Proceedings of the 30th ACM
  International Conference on Multimedia}}. \bibinfo{pages}{6645--6654}.
\newblock


\bibitem[Zhang et~al\mbox{.}(2017)]%
        {zhang2017fast}
\bibfield{author}{\bibinfo{person}{Jing Zhang}, \bibinfo{person}{Yang Cao},
  \bibinfo{person}{Shuai Fang}, \bibinfo{person}{Yu Kang}, {and}
  \bibinfo{person}{Chang Wen~Chen}.} \bibinfo{year}{2017}\natexlab{}.
\newblock \showarticletitle{Fast haze removal for nighttime image using maximum
  reflectance prior}. In \bibinfo{booktitle}{\emph{Proceedings of the IEEE
  Conference on Computer Vision and Pattern Recognition}}.
  \bibinfo{pages}{7418--7426}.
\newblock


\bibitem[Zhang et~al\mbox{.}(2014)]%
        {zhang2014nighttime}
\bibfield{author}{\bibinfo{person}{Jing Zhang}, \bibinfo{person}{Yang Cao},
  {and} \bibinfo{person}{Zengfu Wang}.} \bibinfo{year}{2014}\natexlab{}.
\newblock \showarticletitle{Nighttime haze removal based on a new imaging
  model}. In \bibinfo{booktitle}{\emph{Proceedings of the IEEE International
  Conference on Image Processing}}. IEEE, \bibinfo{pages}{4557--4561}.
\newblock


\bibitem[Zhang et~al\mbox{.}(2020)]%
        {zhang2020nighttime}
\bibfield{author}{\bibinfo{person}{Jing Zhang}, \bibinfo{person}{Yang Cao},
  \bibinfo{person}{Zheng-Jun Zha}, {and} \bibinfo{person}{Dacheng Tao}.}
  \bibinfo{year}{2020}\natexlab{}.
\newblock \showarticletitle{Nighttime dehazing with a synthetic benchmark}. In
  \bibinfo{booktitle}{\emph{Proceedings of the 28th ACM International
  Conference on Multimedia}}. \bibinfo{pages}{2355--2363}.
\newblock


\bibitem[Zhang et~al\mbox{.}(2023)]%
        {Zhang_2023_CVPR}
\bibfield{author}{\bibinfo{person}{Jinghao Zhang}, \bibinfo{person}{Jie Huang},
  \bibinfo{person}{Mingde Yao}, \bibinfo{person}{Zizheng Yang},
  \bibinfo{person}{Hu Yu}, \bibinfo{person}{Man Zhou}, {and}
  \bibinfo{person}{Feng Zhao}.} \bibinfo{year}{2023}\natexlab{}.
\newblock \showarticletitle{Ingredient-oriented multi-degradation learning for
  image restoration}. In \bibinfo{booktitle}{\emph{Proceedings of the IEEE/CVF
  Conference on Computer Vision and Pattern Recognition}}.
  \bibinfo{pages}{5825--5835}.
\newblock


\bibitem[Zhao et~al\mbox{.}(2021)]%
        {zhao2021complementary}
\bibfield{author}{\bibinfo{person}{Dong Zhao}, \bibinfo{person}{Jia Li},
  \bibinfo{person}{Hongyu Li}, {and} \bibinfo{person}{Long Xu}.}
  \bibinfo{year}{2021}\natexlab{}.
\newblock \showarticletitle{Complementary feature enhanced network with vision
  transformer for image dehazing}.
\newblock \bibinfo{journal}{\emph{arXiv preprint arXiv:2109.07100}}
  (\bibinfo{year}{2021}).
\newblock


\bibitem[Zhao(2021)]%
        {zhao2021single}
\bibfield{author}{\bibinfo{person}{Xuan Zhao}.}
  \bibinfo{year}{2021}\natexlab{}.
\newblock \showarticletitle{Single image dehazing using bounded channel
  difference prior}. In \bibinfo{booktitle}{\emph{Proceedings of the IEEE/CVF
  Conference on Computer Vision and Pattern Recognition Workshops}}.
  \bibinfo{pages}{727--735}.
\newblock


\bibitem[Zheng et~al\mbox{.}(2022)]%
        {Zheng2022unsupervised}
\bibfield{author}{\bibinfo{person}{Naishan Zheng}, \bibinfo{person}{Jie Huang},
  \bibinfo{person}{Feng Zhao}, \bibinfo{person}{Xueyang Fu}, {and}
  \bibinfo{person}{Feng Wu}.} \bibinfo{year}{2022}\natexlab{}.
\newblock \showarticletitle{Unsupervised underexposed image enhancement via
  self-illuminated and perceptual guidance}.
\newblock \bibinfo{journal}{\emph{IEEE Transactions on Multimedia}}
  (\bibinfo{year}{2022}), \bibinfo{pages}{1--16}.
\newblock
\urldef\tempurl%
\url{https://doi.org/10.1109/TMM.2022.3193059}
\showDOI{\tempurl}


\bibitem[Zhou et~al\mbox{.}(2022a)]%
        {Zhou2022effective}
\bibfield{author}{\bibinfo{person}{Man Zhou}, \bibinfo{person}{Jie Huang},
  \bibinfo{person}{Xueyang Fu}, \bibinfo{person}{Feng Zhao}, {and}
  \bibinfo{person}{Danfeng Hong}.} \bibinfo{year}{2022}\natexlab{a}.
\newblock \showarticletitle{Effective pan-sharpening by multiscale invertible
  neural network and heterogeneous task distilling}.
\newblock \bibinfo{journal}{\emph{IEEE Transactions on Geoscience and Remote
  Sensing}}  \bibinfo{volume}{60} (\bibinfo{year}{2022}),
  \bibinfo{pages}{1--14}.
\newblock
\urldef\tempurl%
\url{https://doi.org/10.1109/TGRS.2022.3199210}
\showDOI{\tempurl}


\bibitem[Zhou et~al\mbox{.}(2022b)]%
        {AdaptivePan2022MM}
\bibfield{author}{\bibinfo{person}{Man Zhou}, \bibinfo{person}{Jie Huang},
  \bibinfo{person}{Chongyi Li}, \bibinfo{person}{Hu Yu}, \bibinfo{person}{Keyu
  Yan}, \bibinfo{person}{Naishan Zheng}, {and} \bibinfo{person}{Feng Zhao}.}
  \bibinfo{year}{2022}\natexlab{b}.
\newblock \showarticletitle{Adaptively learning low-high frequency information
  integration for pan-sharpening}. In \bibinfo{booktitle}{\emph{Proceedings of
  the 30th ACM International Conference on Multimedia}}.
  \bibinfo{pages}{3375--3384}.
\newblock


\bibitem[Zhou et~al\mbox{.}(2022c)]%
        {NormPan2022MM}
\bibfield{author}{\bibinfo{person}{Man Zhou}, \bibinfo{person}{Jie Huang},
  \bibinfo{person}{Keyu Yan}, \bibinfo{person}{Gang Yang},
  \bibinfo{person}{Aiping Liu}, \bibinfo{person}{Chongyi Li}, {and}
  \bibinfo{person}{Feng Zhao}.} \bibinfo{year}{2022}\natexlab{c}.
\newblock \showarticletitle{Normalization-based feature selection and
  restitution for pan-sharpening}. In \bibinfo{booktitle}{\emph{Proceedings of
  the 30th ACM International Conference on Multimedia}}.
  \bibinfo{pages}{3365--3374}.
\newblock


\bibitem[Zhou et~al\mbox{.}(2022d)]%
        {FourierPan}
\bibfield{author}{\bibinfo{person}{Man Zhou}, \bibinfo{person}{Jie Huang},
  \bibinfo{person}{Keyu Yan}, \bibinfo{person}{Hu Yu}, \bibinfo{person}{Xueyang
  Fu}, \bibinfo{person}{Aiping Liu}, \bibinfo{person}{Xian Wei}, {and}
  \bibinfo{person}{Feng Zhao}.} \bibinfo{year}{2022}\natexlab{d}.
\newblock \showarticletitle{Spatial-frequency domain information integration
  for pan-sharpening}. In \bibinfo{booktitle}{\emph{Proceedings of the European
  Conference on Computer Vision}}. \bibinfo{pages}{274--291}.
\newblock


\bibitem[Zhou et~al\mbox{.}(2022e)]%
        {Zhou_2022_CVPR}
\bibfield{author}{\bibinfo{person}{Man Zhou}, \bibinfo{person}{Keyu Yan},
  \bibinfo{person}{Jie Huang}, \bibinfo{person}{Zihe Yang},
  \bibinfo{person}{Xueyang Fu}, {and} \bibinfo{person}{Feng Zhao}.}
  \bibinfo{year}{2022}\natexlab{e}.
\newblock \showarticletitle{Mutual information-driven pan-sharpening}. In
  \bibinfo{booktitle}{\emph{Proceedings of the IEEE/CVF Conference on Computer
  Vision and Pattern Recognition}}. \bibinfo{pages}{1798--1808}.
\newblock


\bibitem[Zhou et~al\mbox{.}(2022f)]%
        {zhou2022deep}
\bibfield{author}{\bibinfo{person}{Man Zhou}, \bibinfo{person}{Hu Yu},
  \bibinfo{person}{Jie Huang}, \bibinfo{person}{Feng Zhao},
  \bibinfo{person}{Jinwei Gu}, \bibinfo{person}{Chen~Change Loy},
  \bibinfo{person}{Deyu Meng}, {and} \bibinfo{person}{Chongyi Li}.}
  \bibinfo{year}{2022}\natexlab{f}.
\newblock \showarticletitle{Deep fourier up-sampling}. In
  \bibinfo{booktitle}{\emph{Advances in Neural Information Processing
  Systems}}, Vol.~\bibinfo{volume}{35}. \bibinfo{pages}{22995--23008}.
\newblock


\bibitem[Zhu et~al\mbox{.}(2015)]%
        {zhu2015fast}
\bibfield{author}{\bibinfo{person}{Qingsong Zhu}, \bibinfo{person}{Jiaming
  Mai}, {and} \bibinfo{person}{Ling Shao}.} \bibinfo{year}{2015}\natexlab{}.
\newblock \showarticletitle{A fast single image haze removal algorithm using
  color attenuation prior}.
\newblock \bibinfo{journal}{\emph{IEEE Transactions on Image Processing}}
  \bibinfo{volume}{24}, \bibinfo{number}{11} (\bibinfo{year}{2015}),
  \bibinfo{pages}{3522--3533}.
\newblock


\bibitem[Zhu et~al\mbox{.}(2022)]%
        {Zhu_2022_CVPR}
\bibfield{author}{\bibinfo{person}{Yurui Zhu}, \bibinfo{person}{Jie Huang},
  \bibinfo{person}{Xueyang Fu}, \bibinfo{person}{Feng Zhao},
  \bibinfo{person}{Qibin Sun}, {and} \bibinfo{person}{Zheng-Jun Zha}.}
  \bibinfo{year}{2022}\natexlab{}.
\newblock \showarticletitle{Bijective mapping network for shadow removal}. In
  \bibinfo{booktitle}{\emph{Proceedings of the IEEE/CVF Conference on Computer
  Vision and Pattern Recognition}}. \bibinfo{pages}{5627--5636}.
\newblock


\end{thebibliography}











\end{document}